\documentclass[authoryear, preprint,11pt]{elsarticle}
\usepackage{graphicx} 

\usepackage{booktabs} 
\usepackage{multirow} 
\usepackage{makecell} 
\usepackage{amssymb}
\usepackage{amsmath}
\usepackage{ragged2e}
\usepackage{enumerate}
\usepackage{color}
\usepackage{graphicx}
\usepackage{subcaption}
\usepackage{hyperref}
\usepackage{natbib}
\usepackage[dvipsnames]{xcolor}
\usepackage{geometry}
\usepackage[utf8]{inputenc}
\usepackage[english]{babel}
\usepackage{url}

\newcommand{\M}{SMA-Hyper}

\usepackage{algorithm}
\usepackage{algpseudocode}
\algrenewcommand\textproc{\text}
\usepackage{booktabs}
\usepackage{multirow}
\usepackage{enumitem}
\usepackage{balance}
\usepackage{makecell}
\usepackage{threeparttable}
\usepackage{amsmath,amsfonts,mathtools}
\usepackage{mathrsfs}
\usepackage{float}
\usepackage{enumitem}
\usepackage{adjustbox}
\usepackage{tabularx}
\usepackage{ragged2e} 
\usepackage{booktabs}
\usepackage{xcolor}
\usepackage{tabularx}
\usepackage{array}
\usepackage{amsmath}

\newcommand{\redfont}[1]{{\textcolor{red}{#1}}}

\newcolumntype{Y}{>{\centering\arraybackslash}p{3cm}}

\begin{document}
\begin{frontmatter}

\author[inst1]{Xiaowei Gao}
\ead{xiaowei.gao.20@ucl.ac.uk}
\author[inst1]{James Haworth\corref{cor}}
\ead{j.haworth@ucl.ac.uk}

\author[inst1]{Ilya Ilyankou}
\ead{ilya.ilyankou.23@ucl.ac.uk}

\author[inst1]{Xianghui Zhang}
\ead{xianghui.zhang.20@ucl.ac.uk}

\author[inst1]{Tao Cheng}
\ead{tao.cheng@ucl.ac.uk}

\author[inst2]{Stephen Law}
\ead{stephen.law@ucl.ac.uk}

\author[inst3]{Huanfa Chen}
\ead{huanfa.chen@ucl.ac.uk}

\cortext[cor]{Corresponding author}
\affiliation[inst1]{organization={SpaceTimeLab, University College London (UCL)},
            state={London},
            country={UK}}

\affiliation[inst2]{organization={Department of Geography, University College London (UCL) },
            state={London},
            country={UK}}

\affiliation[inst3]{organization={The Bartlett Centre for Advanced Spatial Analysis, University College London (UCL) },
            state={London},
            country={UK}}

\title{~\M: Spatiotemporal Multi-View Fusion Hypergraph Learning for Traffic Accident Prediction}

\newpage

\begin{abstract}

Predicting traffic accidents is the key to sustainable city management, which requires effective address of the dynamic and complex spatiotemporal characteristics of cities. Current data-driven models often struggle with data sparsity and typically overlook the integration of diverse urban data sources and the high-order dependencies within them. Additionally, they frequently rely on predefined topologies or weights, limiting their adaptability in spatiotemporal predictions. To address these issues, we introduce the Spatiotemporal Multiview Adaptive HyperGraph Learning (SMA-Hyper) model, a dynamic deep learning framework designed for traffic accident prediction. Building on previous research, this innovative model incorporates dual adaptive spatiotemporal graph learning mechanisms that enable high-order cross-regional learning through hypergraphs and dynamic adaptation to evolving urban data. It also utilises  contrastive learning to enhance global and local data representations in sparse datasets and employs an advance attention mechanism to fuse multiple views of accident data and urban functional features, thereby enriching the contextual understanding of risk factors. Extensive testing on the London traffic accident dataset demonstrates that the SMA-Hyper model significantly outperforms baseline models across various temporal horizons and multistep outputs, affirming the effectiveness of its multiview fusion and adaptive learning strategies. The interpretability of the results further underscores its potential to improve urban traffic management and safety by leveraging complex spatiotemporal urban data, offering a scalable framework adaptable to diverse urban environments.

\end{abstract}


\end{frontmatter}

\section{Introduction}

Urban traffic management is increasingly challenged by the rapid growth in vehicle numbers and the expansion of road networks, driven by urban economic development and changes in transportation systems. This surge in traffic activity increases the risk of exposure to road accidents, which remain a major public health concern. According to the latest report by the World Health Organisation (WHO), road traffic accidents result in approximately 1.19 million deaths annually worldwide, disproportionately affecting vulnerable road users \citep{WHO2023}. Furthermore, data from the Department for Transport in the UK reveal that despite reduced mobility during the COVID-19 pandemic, the pattern of road injuries from 2018 to 2022 has shown little variation, with such injuries costing nearly 1.5\% of the annual GDP \citep{DfT2023}. These accidents not only compromise public safety but also affect the sustainability and efficiency of urban environments and transportation systems, underscoring the critical need for improved predictive methodologies to prevent such incidents \citep{unece2019}.

The increasing complexity of urban traffic management requires the development of precise models for predicting traffic accidents, crucial for implementing proactive measures that improve both safety and efficiency. Traditionally, accident prediction has predominantly employed time series methods such as Autoregressive Integrated Moving Average (ARIMA) \citep{ihueze2018road}, Gated Recurrent Unit (GRU) \citep{zhang2020modeling}, and Long Short-Term Memory (LSTM) networks \citep{ren2018deep,yuan2018hetero}. Although effective in capturing temporal dynamics, these approaches often overlook significant geographical factors, thus restricting their ability to fully comprehend the intricacies required for accurate predictions. Recent advancements have shifted focus towards integrating spatial relationships, with Graph Neural Networks (GNNs) proving particularly effective in urban settings. Urban environments, with their complex interconnected layouts, are naturally represented as graphs, where nodes signify geographic locations, and edges represent spatial relationships between them. This allows Graph Convolutional Networks (GCNs) \citep{trirat2023mg} and spatiotemporal graph neural networks (ST-GNNs) \citep{zhou2020riskoracle,karim2022dynamic,rahmani2023graph} to model intricate spatial relations essential for understanding traffic dynamics. ST-GNNs, renowned for their ability to uncover nonlinear spatiotemporal relationships, significantly improve the understanding of multiviews contributing to traffic accidents. Their ability to incorporate heterogeneous urban data dimensions further facilitates big data-driven modelling solutions \citep{zhang2020graph}.

\begin{figure}[!htbp]
    \vspace{-0.5mm}
    \begin{minipage}{0.51\columnwidth}
    \centering
    \includegraphics[width=0.80\columnwidth]{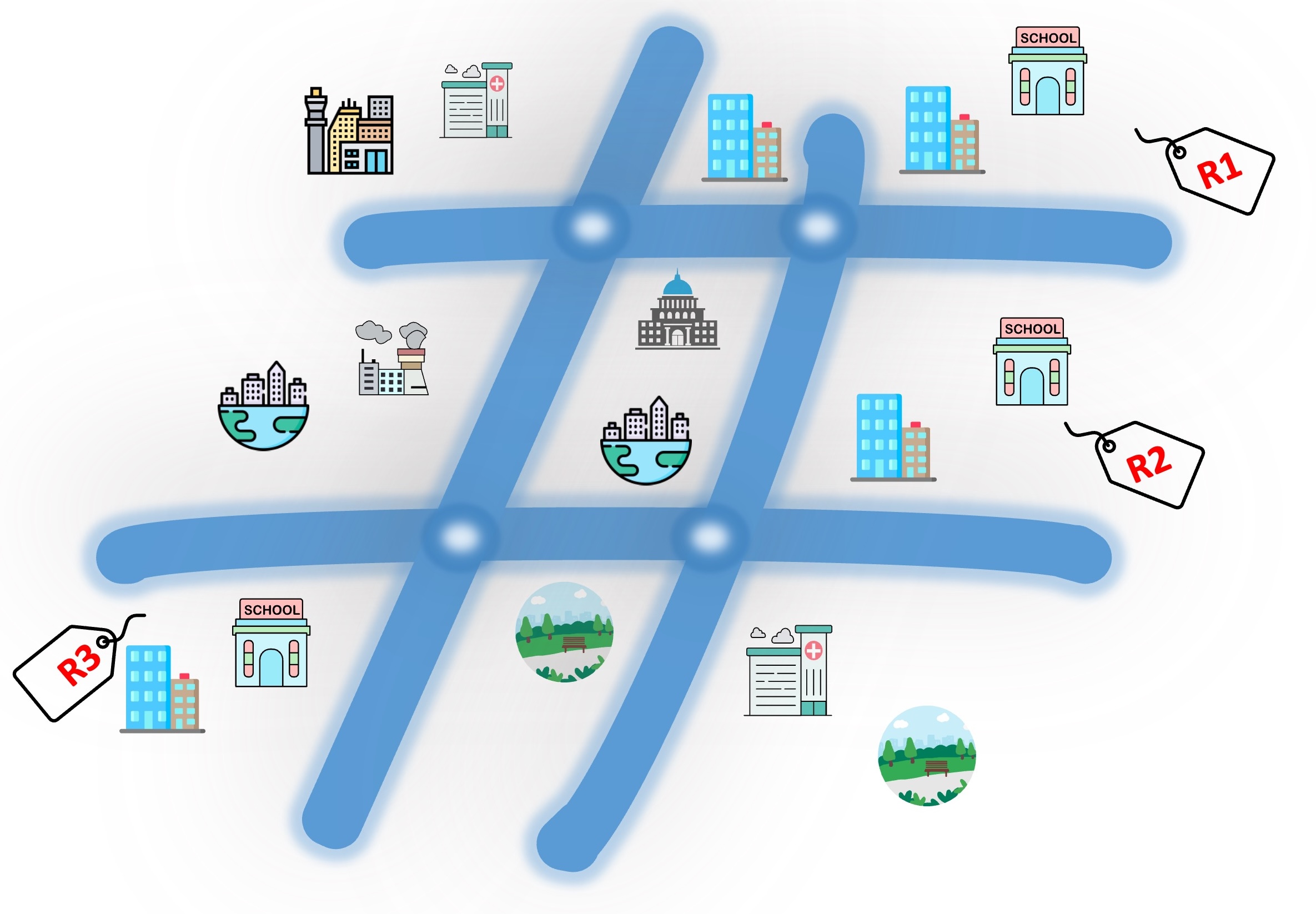}
    \subcaption{Illustration of complex geographic relationships: Regions 1, 2, and 3 share similar POI distributions, yet only Regions 1 and 2 are directly linked by spatiotemporal correlations.}
    \label{fig:1a}
    \end{minipage}
    \begin{minipage}{0.5\columnwidth}
    \centering
    \includegraphics[width=0.99\columnwidth]{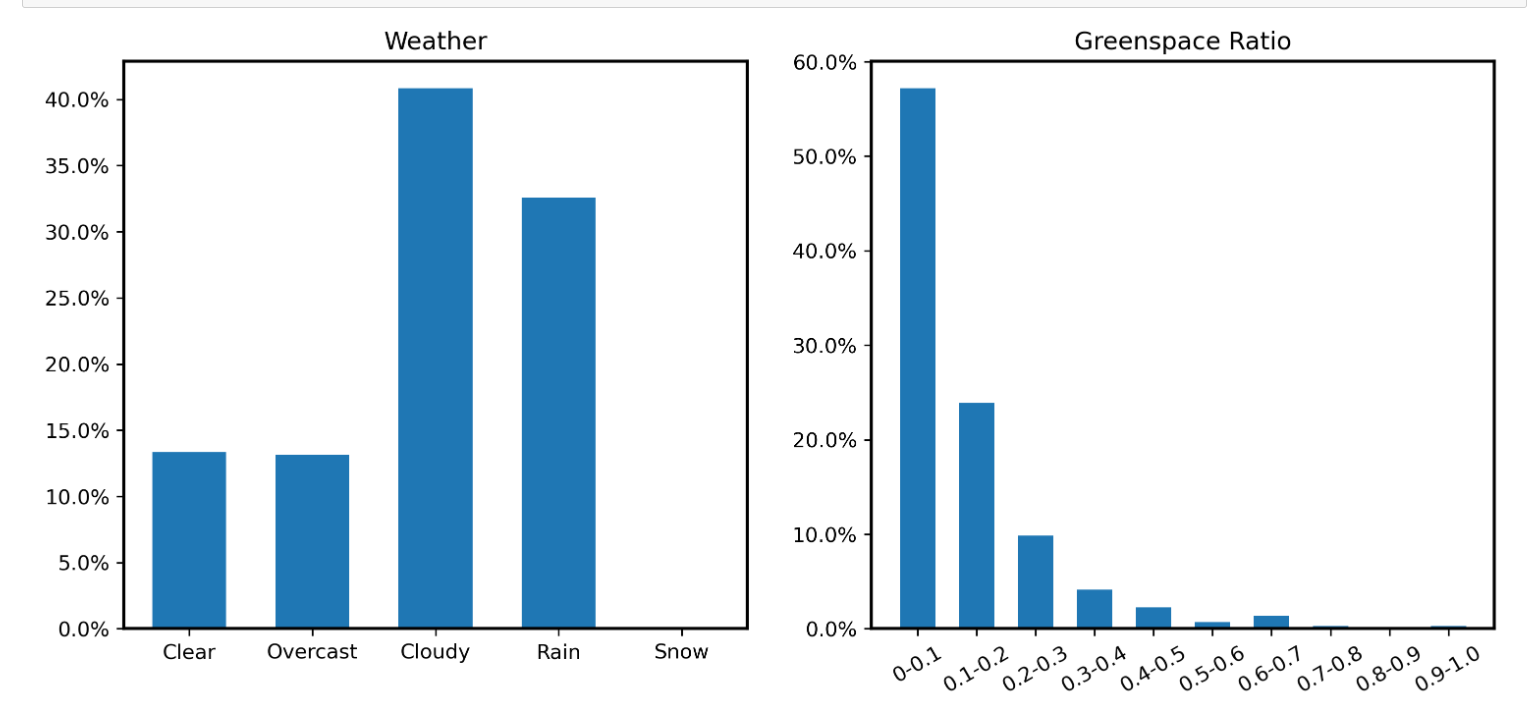}
    \subcaption{The distribution of accident risk ratios in different conditions. The weather conditions include five distinct scenarios, while the greenspace ratio is divided into ten discrete clusters.}
    \label{fig:1b}
    \end{minipage}
    \caption{Case studies illustrating traffic accidents: (a) Geographic relationships among regions with similar POI distributions and their spatiotemporal correlations. (b) Accident risk ratios under varying weather conditions and greenspace distributions.}
    \label{fig:1}
\end{figure}

Despite these technological advances, the common issue in the prediction of traffic accidents is the sparsity in both spatial and temporal dimensions, with incidents that occur nonperiodically in a few locations at particular times. Advanced data enhancement strategies, such as Prior Knowledge-based Data Enhancement (PKDE) proposed by \citet{zhou2020foresee}, have proven to be a promising direction in enriching the data discrimination to better meet the model requirements for effective local neighbourhood information aggregation. Furthermore, the adoption of graph attention techniques uses attention scores to refine node aggregation processes, improving the detection of relationships among localised similar nodes \citep{zhou2020riskoracle,yu2021deep}. Nevertheless, despite the well-addressed sparsity, there are still several unresolved challenges in traffic accident prediction that necessitate further research to enhance the effectiveness of predictive models for urban traffic management:

\begin{itemize}
\item \textbf{Inadequacy of Conventional Pairwise Relations:} Traditional graph models in urban analytics primarily focus on pairwise interactions, limiting analyses to localised relationships between pairs of nodes.  Geographic correlations from points of interest (POIs), as shown in Figure \ref{fig:1a}, suggest that relationships between urban areas go beyond mere proximity, involving complex global similarities across multiple regions, further indicating that the signals of nodes should be propagating within a set of similar attributes distribution nodes. Cities are entangled complex systems whose characteristics intertwine and overlap. Standard graph models, which prioritize immediate node-to-node interactions, fail to capture these extended and high-order spatial dynamics crucial for comprehensive traffic accident analysis. 

\item \textbf{Overlooked Multi-Perspective Urban Data:} Urban environments are rich with diverse data sources, yet current methodologies in traffic accident prediction often fail to utilize this multiplicity effectively. As depicted in Figure \ref{fig:1b}, various external features such as weather conditions and land use types offer unique insights that are pivotal for predicting accident risk. Most existing models rely on simplified single-view graph matrices from accident data itself, which do not adequately account for the multifaceted nature of urban data  \citep{zhou2020foresee,cui2024advancing}. This simplification leads to a lack of consideration in spatiotemporal latent interactions between different types of urban data for the models, ultimately resulting in less convincing predictions.
    
\item \textbf{Static Models in Dynamic Urban Contexts:}  Existing modelling approaches tend to use static affinity matrices for a fixed graph structure based on representations of geographic proximity and attribute similarity, often guided by Tobler’s first and third laws of geography. These predefined model graph structures with strong subjective assumptions make it hard to represent the evolving nature of urban data as well as a lack of understanding of coupled spatiotemporal features, like temporal interaction, configuration similarity, and spatial distance. Furthermore, the rigidity of these models prevents the effective integration of new data or adjustment to multiview data variations, which is crucial for maintaining the generalisation of accident predictions in dynamic urban environments.

\end{itemize}

In response to data-driven challenges in urban traffic management, we introduce the Spatiotemporal Multiview Adaptive HyperGraph Learning (SMA-Hyper) framework. This novel framework enhances the representation and interrelation of urban data within graph models by integrating adaptive graph and hypergraph architectures, tailored to each specific view. By enabling dynamic representations of graph nodes and structures, SMA-Hyper adeptly manages the global latent spatiotemporal dependencies stemming from cross-regional heterogeneous urban data. Further, the SMA-Hyper framework incorporates an attention mechanism to effectively fuse multilayered urban data, such as road conditions and points of interest. This step not only captures the heterogeneity and homogeneity within regional semantic attributes, but also incorporates complex factors that influence traffic accidents. Additionally, SMA-Hyper employs a contrastive learning technique across its graph and hypergraph components to bolster self-supervised learning capabilities, particularly valuable in scenarios characterised by sparse accident data. ~\M maps intricate pairwise and setwise relationships and assembles a comprehensive information transmission architecture. This architecture integrates spatiotemporal traffic accident risk patterns with the latent representations of coupled geographic elements.  Key contributions of the SMA-Hyper framework are summarised as follows.
\begin{itemize}
    \item SMA-Hyper introduces a novel approach to traffic accident prediction by integrating an adaptive hypergraph and graph architecture. This integration enables dynamic data representation and the exploration of complex, higher-order regional dependencies. It also marks a comprehensive consideration from traditional models by utilising an attention mechanism for multiview data fusion and a contrastive learning approach to enhance model learning under data sparsity.

    \item Comprehensive comparison testing on diverse temporal resolutions within the real-world London traffic accident dataset elucidates the framework’s effectiveness, particularly under extremely sparse non-zero instances (3.07\%). The empirical findings underscore a substantial enhancement in accident risk prediction accuracy by 50.71\% - 58.30\% in RMSE and MAE, while maintaining robust recall scores around 40\%, demonstrating the framework’s operational superiority over existing methods.

    \item Extensive supplementary testing, including ablation studies and sensitivity analysis, assesses the impact of each component of SMA-Hyper. These analyses confirm the value of integrating hypergraph learning for capturing spatiotemporal dependencies and the effectiveness of the multiview fusion strategy in improving prediction accuracy. The findings from these studies provide robust evidence of the framework’s adaptability and efficacy in urban traffic settings.
\end{itemize}

The remainder of this paper is organised as follows. Section \ref{sec:lr} reviews related works on deep learning methods for accident prediction and hypergraph learning in urban computing. Section \ref{sec:meth} details the preliminaries, defines the problem, and describes the proposed Adaptive HyperGraph Learning Adaptive Multiview (SMA-Hyper) framework. Section \ref{sec:exp} discusses the experimental setup and results, including ablation and sensitivity analyses. Finally, Section \ref{sec:con} concludes the paper by summarising our findings and suggesting directions for future research.

\section{Literature Review}
\label{sec:lr}

\subsection{Deep Learning for traffic accident prediction}
Early efforts in traffic accident prediction used mainly statistical models or traditional machine learning techniques, which often struggled with the complexity and heterogeneity of large datasets and overlooked crucial spatiotemporal correlations \citep{ali2024advances}. With the advent of advanced artificial intelligence technologies, a shift toward deep learning methods has occurred, leveraging their capacity to unravel complex non-linear spatiotemporal relationships within traffic accident data.

Recurrent Neural Networks (RNNs), especially Long Short-Term Memory (LSTM) networks, have been widely adopted for their efficacy in capturing temporal dynamics. In contrast, convolutional neural networks (CNNs) have been applied to spatial analysis. For instance, \citet{ren2018deep} pioneered the use of LSTM to analyse temporal sequences of traffic accidents, but limited their spatial analysis to basic fully connected layers. Addressing this shortfall, \citet{yuan2018hetero} introduced the Heterogeneous Convolutional LSTM (Heter-ConvLSTM), integrating various urban and external characteristics to better delineate accident risks, albeit within constrained scenarios like their Iowa case study. Further developments led to \citet{moosavi2019accident} integrating LSTM with a latent training representation of POI data and historical traffic event descriptions to capture spatial heterogeneities more effectively. \citet{najjar2017combining} were pioneers in using CNNs to process spatial data derived from satellite imagery to pinpoint accident locations, though their model lacked temporal integration.

Despite the progress, traditional CNNs and RNNs have limitations in capturing extensive spatiotemporal interactions, often restricting their effectiveness to local information learning without broader contextual understanding \citep{cui2024advancing}. Graph Neural Networks (GNNs) have subsequently emerged as a robust solution, adept at encapsulating both localised and expansive spatiotemporal dependencies through the inherent connectivity of urban frameworks. This comprehensive approach enhances the understanding of traffic patterns and facilitates detailed city-wide predictions. For instance, \citet{zhou2020riskoracle} utilised a Differential Time-varying Graph Neural Network (DTGN), complemented by a knowledge-based data enhancement strategy, to address zero-inflated data issues. They, along with \citet{zhou2020foresee} and \citet{wang2021traffic}, implemented multigranularity risk prediction models that define regions with varying grid sizes, proposing to simultaneously predict at different scales. However, this approach has raised feasibility concerns due to the complexity of managing predictions across multiple spatial resolutions effectively. Moreover, recent studies like those by \citet{trirat2023mg} and \citet{yu2021deep} have opted for single-grid size approaches. \citet{yu2021deep} employed spatial convolution layers combined with temporal convolution layers to achieve sophisticated spatiotemporal representations, whereas \citet{trirat2023mg} leveraged a multiview fusion mechanism with attention mechanisms and a Transformer to learn temporal dependencies. Each approach highlights distinct advantages and challenges in capturing urban dynamics but indicates that relying on a single grid size may offer more consistency in model performance across various urban configurations.

However, all current GNN models are based on predefined matrices and manually selected grids, which is limited in adapting to the evolving urban environment and accurately predict traffic accidents in different contexts.

\subsection{HyperGraph Learning Methods in Urban Prediction}
Traditional graph-based models often rely on localised pairwise interactions between nodes, which can restrict the scope of information propagation and fail to capture complex multi-node relationships inherent in urban data \citep{wang2022multitask}. To overcome these limitations, hypergraph learning has been proposed as a powerful alternative. Hypergraphs extend beyond simple pairwise interactions by incorporating multiple nodes into a single hyperedge, thereby facilitating the collaborative extraction of non-Euclidean latent representations from complex data structures \citep{antelmi2023survey, wu2024temporal}. In the realm of urban computing, hypergraph techniques have shown substantial success in various applications, including prediction of mobility flow \citep{alvarez2021evolutionary,wang2024hypergraph}, destination recommendation \citep{yan2023spatio}, and analysis of social interactions \citep{hong2022group}. Each of these applications benefits from the hypergraph’s ability to integrate heterogeneous urban elements into a comprehensive, high-dimensional nonlinear space. This analytical framework improves the representation of urban dynamics, providing a more complete understanding of complex interactions within urban systems.

Despite these successes, the application of hypergraph learning to urban event prediction, such as traffic accidents or crime, remains relatively nascent. Recent studies have begun to explore this domain; for instance, \citet{xia2021spatial} pioneered the use of global cross-regional hypergraphs to predict the occurrence of various categories of crime. In their approach, hyperedges act as intermediate information centres that correlate not only locally connected but also distant regions, thus improving the relational learning capabilities of crime data. Building on this, \citet{li2022spatial} introduced a contrastive learning module to discern spatiotemporal crime patterns at the local and global levels while addressing the challenge of sparse crime data by increasing region self-discrimination. Extending this innovative approach to the prediction of traffic accidents, \citet{cui2024advancing} directly applied the above self-supervised hypergraph learning framework to the prediction of NYC and London traffic accidents. However, their works overlook the integration of urban configurations, focussing predominantly on accident data information. 

\section{Methodology}
\label{sec:meth}
\subsection{Preliminary}

\begin{enumerate}[label=(\roman*)]
    \item \textit{Pairwise Urban Graph}: Under this graph structure, the signals of nodes are transmitted through edges toward another node.  As shown in Figure \ref{fig:2a}, an undirected graph $G(\mathcal{V}, \mathcal{E})$ represents the spatial connectivity between urban regions, with the city divided into $N$ subregions. Here, $\mathcal{V} \in \mathcal{R}^{N}$ denotes the set of vertices, each corresponding to a subregion, and $\mathcal{E} \in \mathcal{R}^{N \times N}$ represents the set of edges, indicating connections between pairs of subregions. The adjacency matrix $A \in \mathcal{R}^{N \times N}$ characterises the spatial relationships between subregions, defined as $A_{i,j} = 1$ if subregions $i$ and $j$ are connected, and $A_{i,j} = 0$ otherwise. 
    
    \item \textit{Setwise Urban Hypergraph}: Under this graph structure, the signals of nodes are propagated within hyperedges toward a cluster of nodes. As shown in Figure \ref{fig:2b}, a hypergraph is defined as $\widehat{G}(\widehat{V}, \widehat{E}, \widehat{W})$, where $\widehat{V} \in \mathcal{R}^{N}$ represents the nodes (subregions), $\widehat{E} \in \mathcal{R}^{N \times I}$ denotes the set of hyperedges capable of linking multiple nodes simultaneously, and $\widehat{W}$ corresponds to the weights of these hyperedges.  A hyperedge captures higher-order spatial dependencies, allowing advanced representations for the global relationships through hypergraph information propagations among each region and corresponding hyperedges.

\begin{figure}[!htbp]
    \vspace{-0.5mm}
    \begin{minipage}{0.51\columnwidth}
    \centering
    \includegraphics[width=0.60\columnwidth]{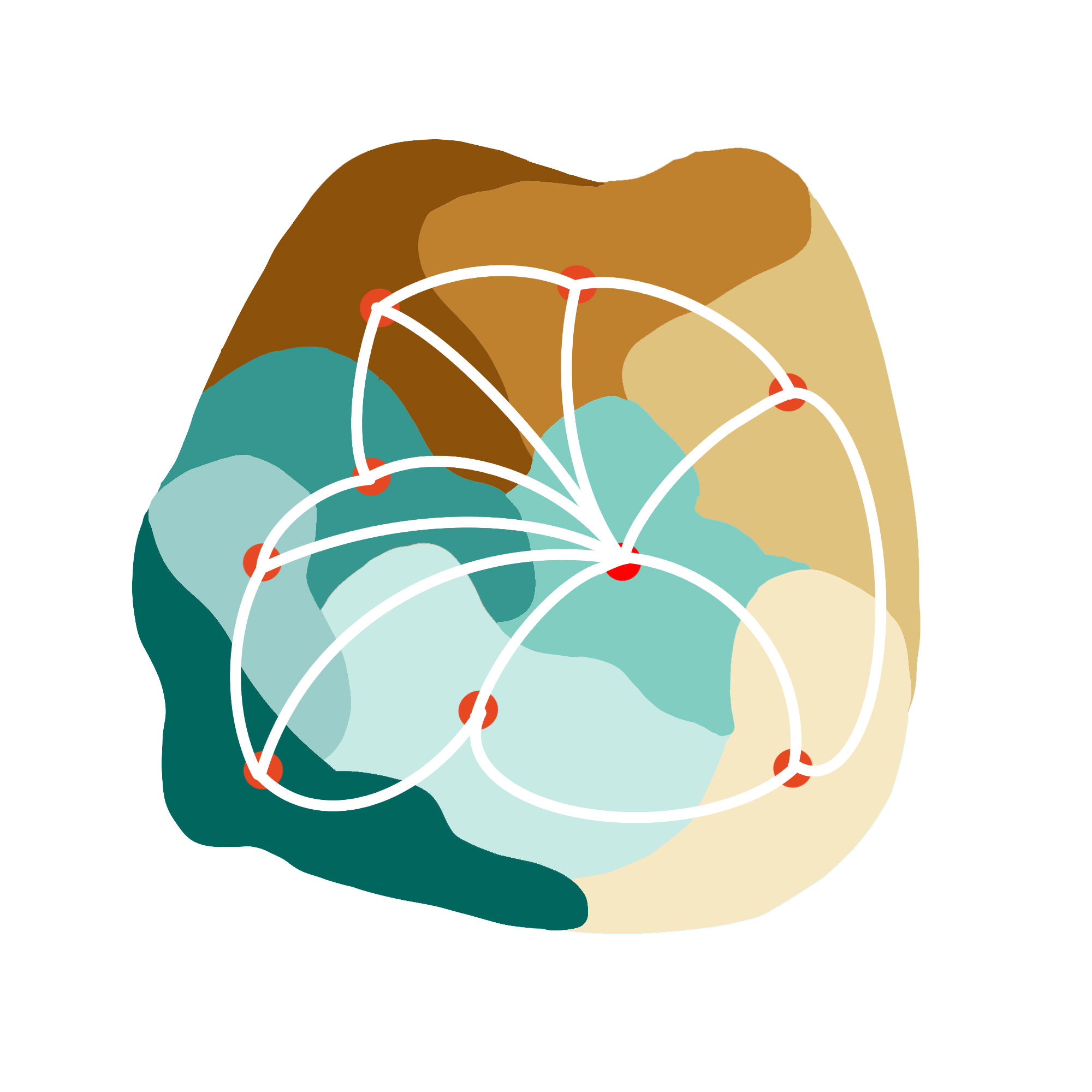}
    \subcaption{Pairwise Graph Structure}
    \label{fig:2a}
    \end{minipage}
    \begin{minipage}{0.5\columnwidth}
    \centering
    \includegraphics[width=0.60\columnwidth]{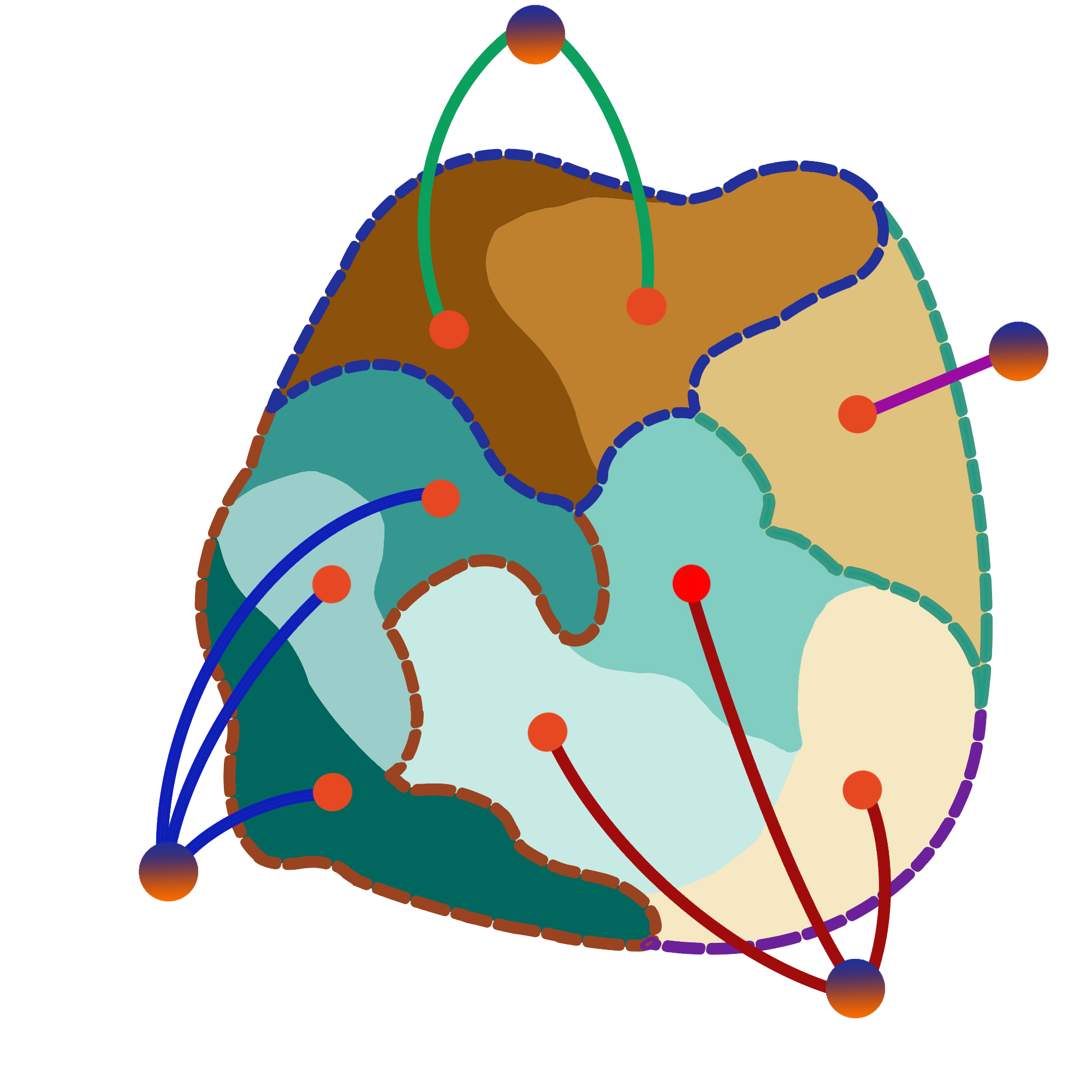}
    \subcaption{Setwise Hypergraph Structure}
    \label{fig:2b}
    \end{minipage}
    \caption{(a) Pairwise graph mapped by pair-pair region spatial connectivity. (b) Setwise Hypergraph focusing on high-order relationships with multihop connections.}
    \label{fig:2}
\end{figure}

    \item \textit{Adaptive Graph Matrix}:  We use a learnable adaptive affinity matrix to represent various non-Euclidean relationships between subregions, offering topological structures for the propagation of node features. The matrix $\mathscr{A}^s \in \mathcal{R}^{N \times N}$ describes the adaptive graph for pairwise-based graph construction, while $\mathscr{H}^s\in \mathcal{R}^{N \times I}$ represents the adaptive hypergraph for setwise-based graph construction.  In this context, $I$ denotes the number of hyperedges.

    \item \textit{Traffic Accident Risk Score}: The accident records are geo-located and time-stamped data with identification of severity. The risk score of traffic accidents combines the severity of heterogeneous levels of accidents within a specific time interval $t$ in individual subregions $n$ as $y_{n,t}=\sum_{i=1}^3f_{n,t}^i\times i$, where $f_{n,t}^i$ denotes the frequency of accidents at severity level $i$ in subregion $n$ during time interval $t$. The weight of severity levels ranges from 1 (slight injuries) to 3 (death), integrating the impacts of varying accident severities to provide a comprehensive risk profile for the subregion.

   \item \textit{Urban Features:}  The set of urban features is represented by two views: points of interest (POI) and road statistical information. The POI distribution includes the density of different types of POI categories, while road information encompasses specifics such as road type, average width, and infrastructure details. These features are represented as $X^P \in \mathcal{R}^{N \times d_P}$ (POI) and $X^R \in \mathcal{R}^{N \times d_R}$ (Road), respectively, in which $d_P$ and $d_R$ are the dimension of the corresponding feature embeddings.

   \item \textit{External Features}: External features refer to the influences other than the urban geographical environment that affect accidents. These factors mainly include meteorology and calendar information. Meteorology features include temperature, snow depth, and visibility, while calendar information covers holidays, peak travel times, and days of the week. These features are represented as $X^{M} \in \mathcal{R}^{N \times T \times {d_M}}$ (Meteorology) and $X^{C} \in \mathcal{R}^{N \times T \times {d_C}}$ (Calendar), where $d_M$ and $d_C$ indicate the dimension of feature embedding, respectively.

\end{enumerate}

\subsection{Problem Definition}

Given the adaptive graph and also the adaptive hypergraph $\mathbf{G} \in (G, \widehat{G})$ of the city, the \textbf{input} is the historical accident risk series $\mathbf{X_t}$ for $t = 1, 2, \ldots, T$, the \textbf{output} is to predict the accident risk score, resulting from $y_{n,t}$, for each subregion in the next time intervals $\tau$. Formally, the problem can be formulated as finding a function $P_{\Theta}(\cdot)$ such that:
\begin{equation}
[\mathbf{X_t}; \mathbf{G}] \xrightarrow{P_{\Theta}(\cdot)} [Y_{t+1}, \ldots, Y_{t+\tau}]
\end{equation}
where $P_{\Theta}(\cdot): \mathbb{R}^{ N \times T} \rightarrow \mathbb{R}^{N \times \tau}$ is the target function mapping from spatiotemporal features of the past $T$ steps to the predicted accident risk in the next $\tau$ steps. $\Theta$ represents the learnable parameters and $Y_t \in \mathbb{R}^N$ is the accident risk score metrics for all nodes in the time step $t$.

\subsection{Tackling Sparse Accident Data}

A significant challenge in the prediction of fine-grained traffic accidents is zero inflation. To address this, we use the PKDE method to augment zero-value data, which was proposed by \citet{zhou2020foresee,zhou2020riskoracle}. This approach adjusts zero-value elements based on statistical accident information across different subregions, incorporating prior knowledge by assigning negative lower risk scores to nonaccident instances and positive higher scores to accident instances to improve the model's ability to discriminate accident risk levels. Specifically, we replace the risk of zero value for the subregion $i$ in each time interval with the negative statistical intensity of the accident $\pi_{i}$: $\pi_{i} = b_1 \log_2 \epsilon_{i} + b_2$. 
Here, $\epsilon_{i}$ represents the statistical accident indicator that quantifies the accident risk of subregion $i$. The coefficients $b_1$ and $b_2$ ensure that the range of $\pi_{i}$ reflects the situation of actual risk levels. 

The zero value elements are therefore adjusted based on the proportion of the total risk level of the subregion and assigned to the range $[-1, 0]$ through the logarithmic and linear transformations described above. Nonzero values are normalised to the interval $[0, 1]$, thereby preserving the rank of actual accident risks. It enhances the distinction between positive and negative samples by making the accident intensity value negative and distinct. 

\subsection{Spatiotemporal Multiview Adaptive HyperGraph Learning Framework}

\begin{figure}[!htbp]
    \centering
    \includegraphics[width=0.90\columnwidth]{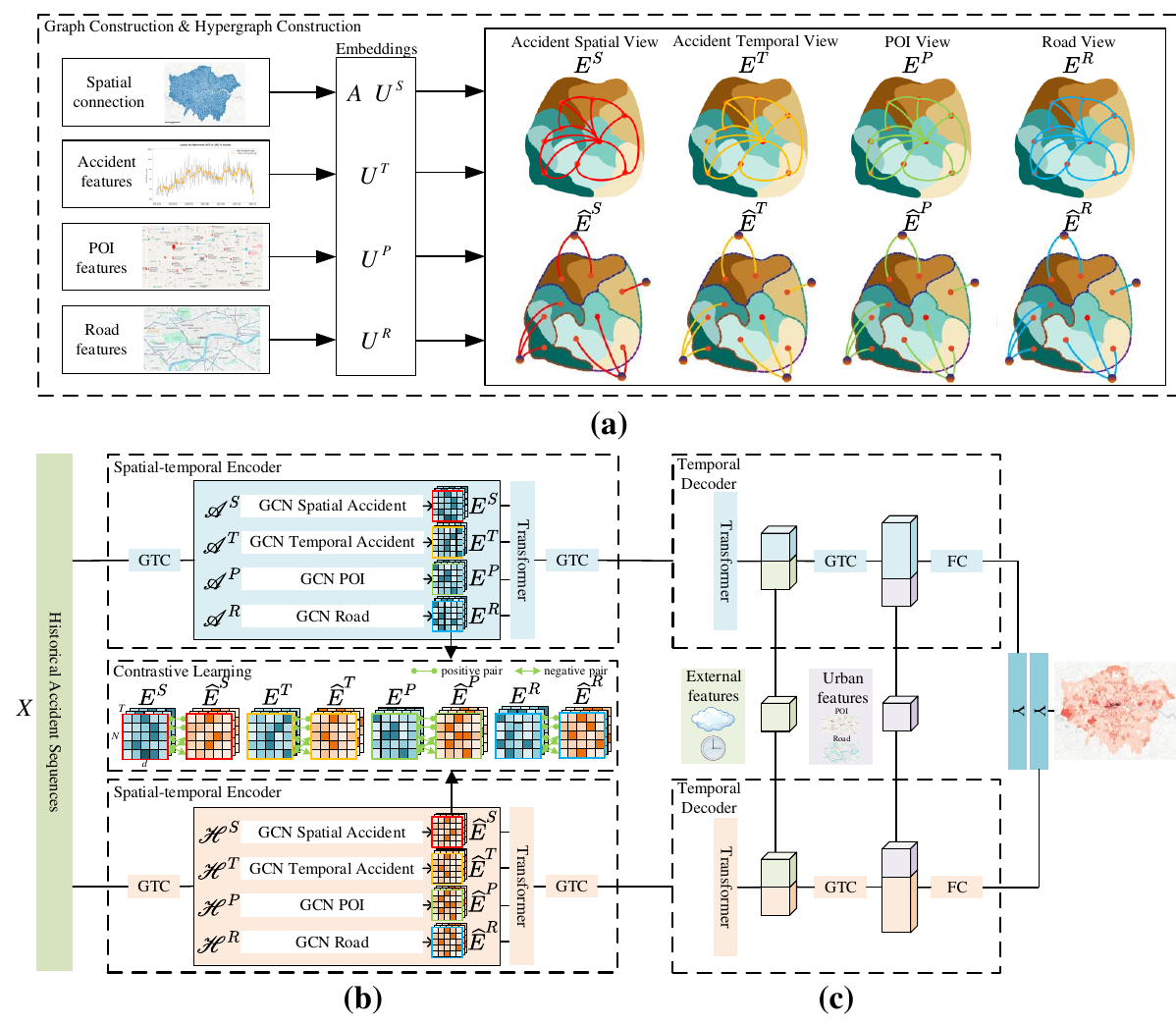}
    \caption{Overview of the Spatiotemporal Multiview Adaptive HyperGraph Learning (SMA-Hyper) Framework. Section (a) depicts the graph construction process, integrating historical accident data with urban features into the graph and hypergraph structures. Section (b) presents the encoder phase, incorporating contrastive learning to enhance feature representation and differentiation. Section (c) illustrates the multi-view fusion decoder, which synthesizes various urban data streams like POIs and road information, culminating in the predictive output for traffic accident risks}
    \label{fig:frame}
\end{figure}

In this section, we describe the details of the proposed SMA-Hyper model, an adaptive framework for predicting traffic accidents, as illustrated in \redfont{Figure \ref{fig:frame}}, it is primarily composed of four main modules:

1) Multiview adaptive graph and hypergraph construction: Graphs and hypergraphs are constructed based on spatial relationships, urban features, and accident sequences, capturing low-order and high-order relationships between subregions, respectively. 2) Multiview spatiotemporal feature encoder: In the multiview context, the propagation of accident characteristics between pairwise edges and setwise hyperedges captures the complex spatiotemporal correlations of accident occurrences. 3) Temporal decoder: Integrating external characteristics, the spatiotemporal embedding vectors of each subregion are decoded through gated convolutions to generate predictions for future accident sequences. 4) Local-global contrastive learning: The contrastive learning between multiview graphs and hypergraphs offers additional self-supervisory signals to the model, mitigating the negative impact of data sparsity on model learning.

Detailed information and calculation processes are outlined below.

\subsubsection{Multiview AdaptiveGraph and Hypergraph Learning}

\textbf{Multiview Graph Embedding} Traditional ST-GNN models often rely on predefined topologies based on spatial connections or interaction weights. This approach lacks comprehensive geographic relationships and depends on a single criterion, which limits its ability to generalize across scenarios involving both static and dynamic urban features \citep{wang2024data, xiong2023adaptive}. Additionally, defining these criteria requires specific spatial knowledge, which is not always available or easily generalizable, especially when multiple types of views are considered. Inspired by Tobler’s first law and the third law of geography, which suggest that closer spatial distances lead to similar target labels and that long-distance spatial interactions can occur with similar urban configurations, we propose adaptive conventional graph structure definitions to incorporate evolving configurations from multiple views rather than relying on a single criterion.

To be specific, for static urban features including POI and the complexity of the road network, we employ fully connected layers to achieve characteristic mapping and denote them as $U^s, s \in \{ P, R \}$. Meanwhile, the similarity of historical accident sequences can also provide effective information for uncovering correlations between different sub-regions. Due to the dynamic nature of the sequences, we utilise a combination of stacked convolutional layers and fully connected layers to generate its embedding\citep{zhang2022dynamic}, and denote it as $U^T$.

After embedding, the non-Euclidean spatiotemporal correlations between sub-regions are mapped through multiple views: accident spatial view, accident temporal view, POI view and road view, with each view is constructed by the pairwise graph $\mathscr{A}^s$ and the setwise hypergraph $\mathscr{H}^s$, where $s$ denotes different views.

\textbf{Pairwise Graph Construction}  Accident spatial view reflects the impact of spatial proximity on accident prediction. According to Tobler's First Law of Geography, two adjacent regions tend to exhibit a closer spatiotemporal correlation. Therefore, we utilize the adjacency matrix $A$ as the topology for this view and denote it as $\mathscr{A}^{S}$. 

For the construction of the remaining three views based on adaptive manner, the extracted features mentioned above are used to establish connections between sub-regions based on the geography's third law \citep{zhu2018spatial}. For instance, two sub-regions with similar POI distribution or historical accident sequences may share similar accident patterns. Hence, the product of features is applied to measure the similarity, after which the activation function $Tanh(\cdot)$ and $Relu(\cdot)$ are used to scale the values to [-1,1] and eliminate unwarranted negative values:

\begin{equation}\mathscr{A}^{s}=Relu(Tanh(U^{s}U^{s\top})),\end{equation}

where $s$ here represents the accident temporal view, POI view and road view, respectively.


For each sub-region, aggregating information from thousands of other sub-regions may introduce a significant amount of noise, thereby reducing predictive performance. Therefore, we set a threshold $k$ for the maximum number of edges related to each sub-region to control the sparsity of graph. Specifically, we first sort the values in each row of matrix $\mathscr{A}^{s}$ and retain only the top $k$ values:

\begin{equation}indices_i=topk(\mathscr{A}^{s}[i,:]), i=1,2,...,N \end{equation}
\begin{equation} \mathscr{A}^{s}[i,:] = \mathscr{A}^{s}[i,indices_i], i=1,2,...,N\end{equation}

Therefore, we can obtain the pairwise adaptive graphs under the four views, which are represented as follows: $\mathscr{A}^{S}$ (accident spatial view), $\mathscr{A}^{T}$ (accident temporal view), $\mathscr{A}^{P}$ (POI view), $\mathscr{A}^{R}$ (road view).

\textbf{Setwise Hypergraph Construction} The same functional distributions may exert similar effects on the accident patterns across multiple sub-regions. Traditional graphs can only model pairwise relationships between two entities, which results in their limitations, and it is extremely difficult to integrate distant information through stacked multi-layer signal transmission. Therefore, we employ hypergraphs to model the high-order relationships among cross-regional nodes from multi-views, which may exhibit similar accident patterns, POI distributions, or road characteristics. Consequently, we adopt an adaptive hypergraph construction approach to automatically mine learnable high-order similarities. 

Different from the adaptive pairwise graph, the adaptive setwise hypergraph is represented by an incidence matrix $\mathscr{H}^{s}\in\mathcal{R}^{N\times I}$, where $I$ denotes the number of hyperedges. Each hyperedge is capable of simultaneously connecting multiple sub-regions, thereby capturing global higher-order dependencies. We derive $\mathscr{H}^{s}$ by multiplying a feature matrix $K^{s}\in\mathcal{R}^{d_s\times I}$, where $d_s$ is the embedding dimensions of embedding set $s$. Furthermore, we utilize $Tanh(\cdot)$ and $Relu(\cdot)$ to confine the values of the correlation matrix within a reasonable range. Similarly, we control the sparsity of hyperedges through the parameter $k$:

\begin{equation}\mathscr{H}^{s}=Relu(Tanh(U^{s}K^{s})),\end{equation}
\begin{equation}indices_i=topk(\mathscr{H}^{s}[i,:]), i=1,2,...,N \end{equation}
\begin{equation} \mathscr{H}^{s}[i,:] = \mathscr{H}^{s}[i,indices_i], i=1,2,...,N\end{equation}

where $s$ here also represents the accident temporal view, POI view and road view, respectively.

It is noteworthy that in accident spatial view, we employ a learnable matrix $U_n^{S}$ to represent the positional embeddings of subregions, with the expectation that the constructed adaptive hypergraph will uncover higher-order positional relationships.  

\begin{equation}\mathscr{H}^{S}=Relu(Tanh(U^{S}K^{S})),\end{equation}

Therefore, we can obtain the corresponding setwise adaptive hypergraphs under the four views, which are represented as follows: $\mathscr{H}^{S}$ (accident spatial view), $\mathscr{H}^{T}$ (accident temporal view), $\mathscr{H}^{P}$ (POI view), $\mathscr{H}^{R}$ (road view).



\subsubsection{Multi-View Spatio-temporal Features Encoder}

The extraction of complex spatiotemporal features based on learned multi-view graphs and hypergraghs presents the next challenge. Graph Convolutional Network (GCN) is a technique widely employed in the field of spatiotemporal data mining, which facilitates the learning of spatial dependencies by aggregating signals from adjacent nodes. In contrast, Heterogeneous Graph Convolutional Network (HGCN) operates based on high-order relationships within hypergraphs, contributing to the generation of a more robust spatial representation. In this context, we propose the multi-view spatio-temporal features encoder illustrated in Figure \ref{fig:frame} (b), which encompasses both graph-based and hypergraph-based encoders. Each encoder module is composed of a "sandwich" structure comprising gated temporal convolution blocks and spatial convolution blocks. Lastly, the spatial features extracted by spatial convolution blocks are integrated through an attention based fusion block.

\textbf{Accident Embedding} Firstly, an embedding layer is applied to the input $X$ to generate the initial representation of the historical accident sequence. Specifically, each element in $X$ is multiplied by a randomly initialized vector $e$ to obtain its representation: $E_{n,t} = X_{n,t} \cdot e$.

\textbf{Gated Temporal Convolution Block} The prediction of accident risks is inherently a time series forecasting problem. hence, we extract temporal features of accident sequences through a dedicated temporal learning block, names GTC (Gated Temporal Convolution) block. Specifically, the temporal learning block consists of two stacked temporal convolution modules, each of which models the contextual information of the sequence via a gating mechanism to enhance the expressiveness. The equation for each temporal convolution module is as follows:

\begin{equation}gt(E_{in})=sigmoid(W_1E_{in}+b_1)\end{equation}
\begin{equation}E_{out}=(1-gT(E_{in}))\cdot E_{in}+gt(E_{in})\cdot (W_2E_{in}+b_2)\end{equation}

where $E_{out}$ represents the output of the time learning block in layer $l$. $gt(\cdot)$ denotes the gating unit, which facilitates the fusion of residual connections and learned features. Lastly, $W_1,W_2$ and $b_1,b_2$ are the learnable parameters of the convolutional layers.

\textbf{Graph Convolution Block} Extensive research in the past has demonstrated that Graph Convolutional Network (GCN) possesses remarkable capabilities in learning and aggregating spatial relationships. Based on the constructed multi-view topology, stacked graph convolutional blocks are capable of capturing spatial relationships ranging from one-hop to multi-hops. The spatial convolution block is defined as follows:

\begin{equation}E_{out}^{s}=ReLU(\mathscr{A}^s E_{in} W_3),\end{equation}

where $\mathscr{A}^s$ is the normalized adjacency matrix in view $s$, and $E_{out}^{s}$ is the hidden representation obtained therefrom. $W_3$ represents the weights of the corresponding convolutional kernels. It is worth noting that graph convolution block generates corresponding representations under the topology of 4 views respectively. Consequently, we can derive multi-view representations: $E^{s}, s \in \{S,T,P,R\}$.

\textbf{Hypergraph Convolution Block} In previous works, the higher-order relationships between sub-regions have been overlooked. Hypergraph convolution can uncover the high-order spatial relationships between sub-regions, thereby achieving more advanced representations. Unlike graph convolution, hypergraph convolution calculates based on the incidence matrix and can be expressed as:

\begin{equation}\widehat{E}_{out}^{s}=ReLU({D_R^{s}}^{-1/2}\mathscr{H}^{s\top}{D_E^{s}}^{-1/2}ReLU({D_E^{s}}^{-1/2}\mathscr{H}^{s}{D_R^{s}}^{-1/2}E_{in}^{s}W_4))\end{equation}

where ${D_R^{s}}$ and ${D_E^{s}}$ denote diagonal degree matrices of $\mathscr{H}^s, s \in \{S,T,P,R\}$. $W^H$ denotes the weights of the hypergraph convolution. Similarly, based on the incidence matrix $\mathscr{H}^s$ constructed under 4 views mentioned above, we can obtain the corresponding hypergraph representations $\widehat{E}^{s}, s \in \{S,T,P,R\}$.

\textbf{Attention-based Multi-View Features Fusion Block:} For different accident sequence patterns, it is necessary to pay closer attention to specific views, where the multi-head attention mechanism can assist in capturing the complex relationships among them. For adaptive pairwise graphs and adaptive setwise hypergraphs, we utilize the Transformer module to fuse the spatial-temporal representations of 4 views, respectively:

\begin{equation}E_{out}=Transformer(E_{in}^{S}, E_{in}^{T}, E_{in}^{P}, E_{in}^{R}),\end{equation}

\begin{equation}\widehat{E}_{out}=Transformer(\widehat{E}_{in}^{S}, \widehat{E}_{in}^{T}, \widehat{E}_{in}^{P}, \widehat{E}_{in}^{R}),\end{equation}

where $E_{out}$ and $\widehat{E}_{out}$ denote the multi-view fused  spatial-temporal representions based on graphs and hypergraphs, respectively.





\textbf{Attention-based Multi-Layer Features Fusion Block:} The accident features are processed sequentially through the gated temporal convolution block, the multi-view graph and hypergraph convolution block, and another gated temporal convolution block, resulting in the outputs of each layer. The outputs of different layers encapsulate spatio-temporal features of varying levels. The higher-level outputs aggregate multi-step spatiotemporal information through the propagation and convergence of stacked signals. Traditional methods based on average pooling assign the same weight to each layer, which may limit the expressiveness of the fused representations. Therefore, in the face of multi-layer spatial-temporal features $E^{l}, l=1,2,..., L$ and $\widehat{E}^{l}, l=1,2,..., L$, we employ an attention module to capture the relationships among the different layers:

\begin{equation}E^{encoder}=Transformer(E_{t}^{1}, E_{t}^{2},..., E_{t}^{L}),\end{equation}

\begin{equation}\widehat{E}^{encoder}=Transformer(\widehat{E}_{t}^{1}, \widehat{E}_{t}^{2},..., \widehat{E}_{t}^{L}),\end{equation}

Here,we can get the graph-based and hypergraph-based encoder outputs: $E^{encoder}$ and $\widehat{E}^{encoder}$.

\subsubsection{ Temporal Decoder Incorporating External Features }

Dynamic external features encompass weather and calendar information. Extremes in weather conditions, such as rain and snow, may result in a further increase in accident risk. Similarly, during specific time periods within a week, heavier traffic congestion may also increase the risk of accidents . Hence, when decoding, dynamic external features should be incorporated as part of the input and fused with encoder outputs, and we can obtain the decoder outputs as follows:

\begin{equation}E^{decoder}=GTC([E^{encoder},W^{M}X^M,W^{C}X^C]),\end{equation}


where $GTC(\cdot)$ indicates the gated temporal convolution block. $[\cdot]$ refers to the concatenation option, in which $X^M, X^C$ are matrices of external features, and $W^{M}, W^{C}$ are learnable parameters of external features embedding. Following this, we can obtain the decoder outputs $E^{decoder}$ and $\widehat{E}^{decoder}$, respectively, based on the outputs of the graph encoder and hypergraph encoder .

Finally, incorporating urban features including POI and road statistical information is crucial for prediction as well. For instance, certain specific POIs such as supermarkets and hospitals, due to their surroundings with higher traffic density, may lead to a higher level of accident risk. Therefore, the last prediction layer can be modeled as:

\begin{equation}Y=FC(ReLU(FC([E^{decoder}, W^{P}X^P, W^{R}X^R])))\end{equation}

In this formula, we recalculate the embeddings for urban features, as the purpose here is to assess the impact of urban features on accident risk prediction. In contrast, embeddings $U^P$ and $U^R$ are designed to uncover the similarities in the distribution of urban features across different regions. Ultimately, the output $Y$ of the prediction layer is taken as the predictive outcome of the model.

\subsubsection{Local-Global Contrastive Learning}

The local spatio-temporal encoder based on pairwise graph is capable of aggregating local information from multiple hops through multi-layer graph convolutions, while the global spatio-temporal encoder based on setwise hypergraph can facilitate higher-order cross-region information learning. To facilitate the alignment and integration of the information of local and global encoders, we employ cross-view contrastive learning. It promotes the collaborative supervision between the local and global encoders through self-supervised learning, thereby providing additional supervisory signals and mitigating the impact of data sparsity. Moreover, the resultant more robust spatio-temporal embeddings can reduce the influence of input noise.

\redfont{Specifically, in our cross-view contrastive learning module, embeddings based on pairwise graphs and embeddings based on setwise hypergraphs from the same view and the same sub-region are treated as positive samples, while those from different views or sub-regions are considered negative samples. Consequently, the loss function can be formulated as follows:}

\begin{equation}\mathcal{L}^{(C)}=\sum_{s}\sum_{l=1}^L\sum_{n=1}^N\log\frac{\exp(\cos(E_{n}^{l,s},\widehat{E}_{n}^{l,s}))}{\sum_{n^{\prime}}\exp(\cos(E_{n}^{l,s},\widehat{E}_{n^{\prime}}^{l,s}))}\end{equation}

In which $s \in \{ S,T,P,R \}$ represents spatial accident view, temporal accident view, POI view and Road view, respectively, and $ L $ is a hyperparameter that denotes the number of stacked layers of multi-View spatio-temporal features encoder.

\subsubsection{Model Optimisation}

In the model optimisation process, we employ the Mean Squared Error (MSE) function to gauge the accuracy of the prediction results. Incorporating the aforementioned contrastive learning loss, the ultimate joint loss function obtained is as follows:

\begin{equation}\mathcal{L}=\sum(Y-\widehat{Y})^2+\lambda_{1} \mathcal{L}^{(C)}+\lambda_{2}\|\boldsymbol{\Theta}\|_{2}^{2}
\end{equation}

where $Y_t$ and $\widehat{Y}_t$ represent the actual and predicted accident risk scores, respectively. $\boldsymbol{\Theta}$ denotes the learnable parameters, and a regularisation loss based on the $L_2$ norm is applied to prevent overfitting. Lastly, $\lambda_{1}$ and $\lambda_{2}$ are hyperparameters, representing the weights for contrastive learning loss and the regularisation loss, respectively.

\section{Experiments}
\label{sec:exp}

\subsection{Dataset}

The proposed SMA-Hyper model was evaluated using London traffic accident data from the STATS19 dataset\footnote{\url{https://www.data.gov.uk/dataset/cb7ae6f0-4be6-4935-9277-47e5ce24a11f/road-safety-data}}, officially published by the Department of Transport, UK. This dataset is collected from police reports, so incidents without injuries are not included. The dataset provides detailed information about the vehicles and casualties involved, as well as the geographical location and time of each incident. The data were pre-processed and formatted using the R package \textit{‘stats19’} by \citet{lovelace2017introducing}.

For this study, we used a total of 23,096 accidents in London from 1 January to 31 December 2021. The breakdown of accident severity is as follows: slight accidents account for 19,624 incidents (84.97\%), serious accidents for 3,399 incidents (14.72\%), and fatal accidents for 73 incidents (0.31\%).

For the spatial unit, we used Middle Super Output Areas (MSOAs), which are official census units with an average area of 19 square kilometers. Unlike manually divided grids, MSOAs offer several advantages for deep learning embeddings: (1) Consistency with Demographic Data: MSOAs are designed to be consistent with demographic data, providing a robust framework for integrating socio-economic variables into the model. This improves the model’s ability to learn complex exposure patterns related to traffic accidents. (2) Population homogeneity: MSOAs contain populations that are relatively homogeneous in terms of socio-economic characteristics. This helps reduce variability within each unit, making the model's learning process more effective. (3) Administrative Relevance: Using MSOAs aligns the model with administrative boundaries, which is beneficial for policy-making and urban planning. Local authorities can directly use predictions made by the model for targeted interventions. (4) Geographic accuracy: MSOAs provide more accurate geographical representations compared to grid-based divisions. They account for natural and built environment features, which are crucial for accurately capturing the spatial dynamics of traffic accidents. 

\begin{figure}[htbp]
    \centering
    
    \begin{subfigure}[b]{\textwidth}
        \centering
        \includegraphics[width=0.7\textwidth]{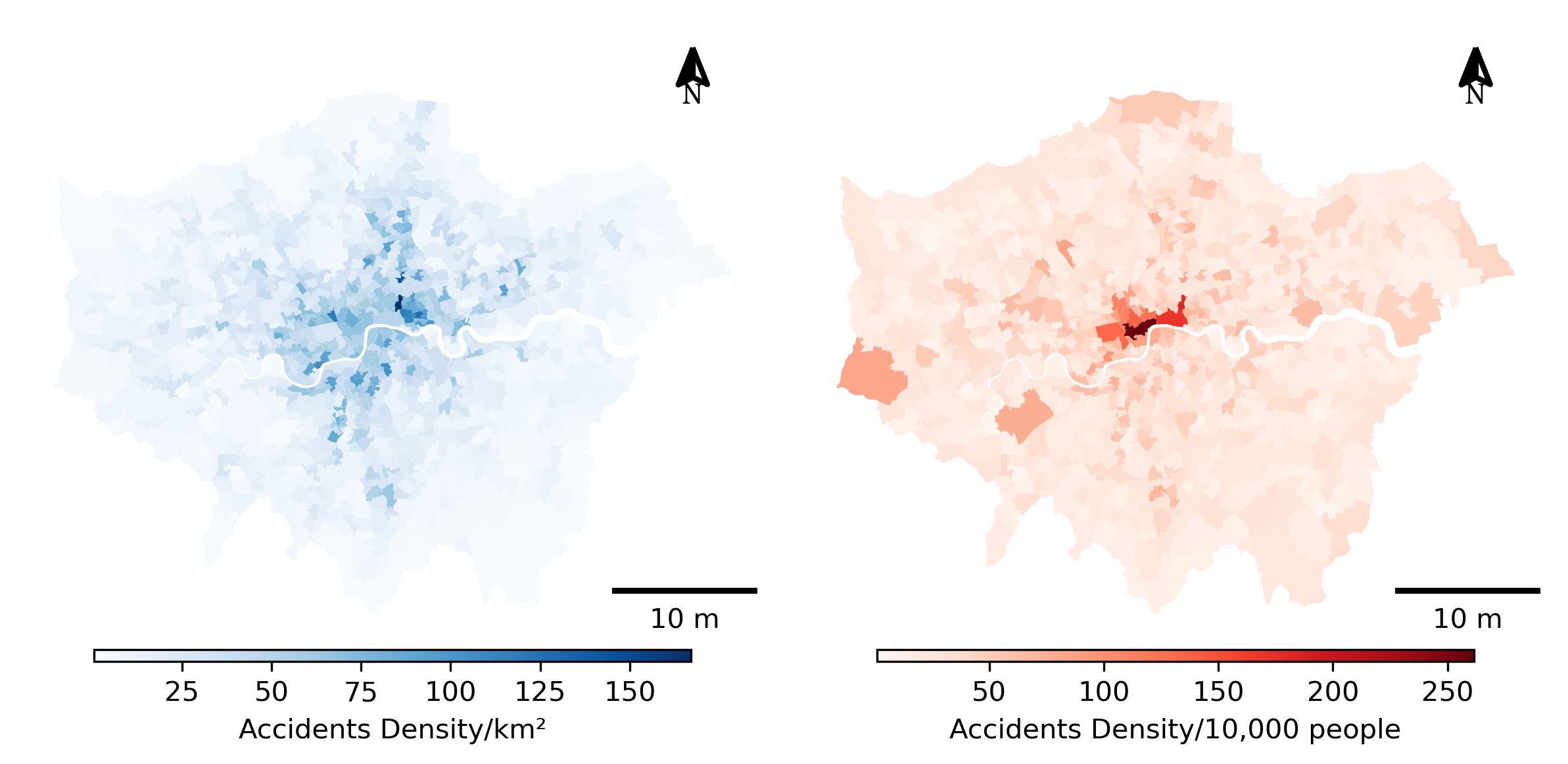}
        \caption{The density map of traffic accidents for MSOAs. It is calculated by dividing the number of non-zero elements (accidents happened) by the size and people count of each MSOA.}
        \label{fig:acc}
    \end{subfigure}
    \hfill
    \begin{subfigure}[b]{\textwidth}
        \centering
        \includegraphics[width=0.7\textwidth]{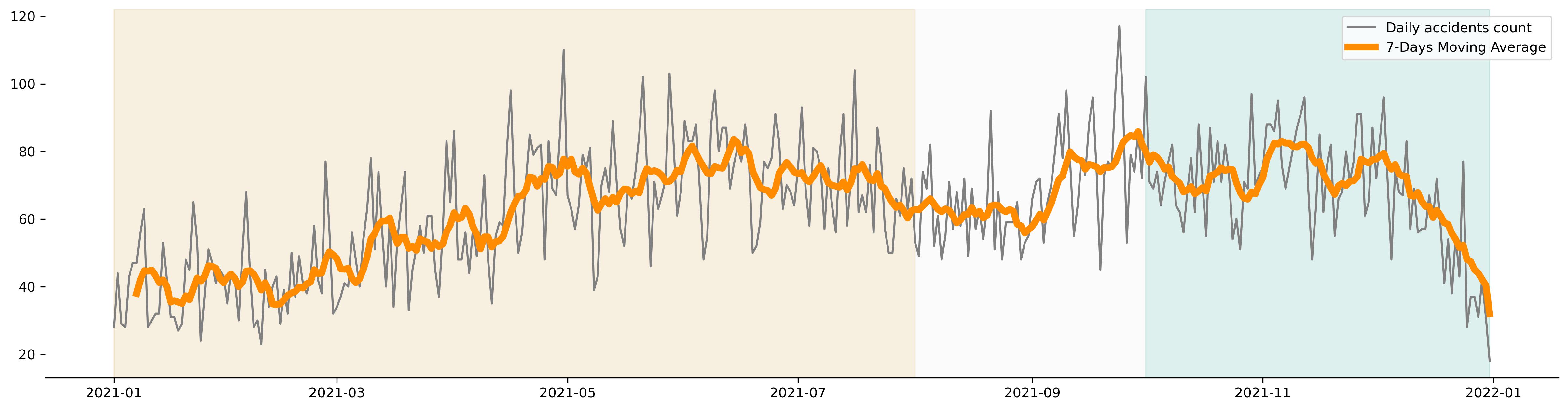}
        \caption{Daily cycling accident counts and 7-day moving average from January 1, 2021, to December 31, 2021. The grey line shows daily accident counts, and the orange line represents the 7-day moving average. Shaded regions indicate different periods: training (light yellow, January 1 - August 1), validation (white, August 1 - October 1), and testing (light green, October 1 - December 31). This segmentation aids in model training and evaluation.}
        \label{fig:daily_accidents}
    \end{subfigure}
    
    \caption{Spatial and Temporal Comparison of traffic accident data visualizations of London, 2021.}
    \label{fig:comparison}
\end{figure}

Therefore, we spatially joined the traffic accident data points to a total of 1,002 MSOAs, with temporal prediction resolutions of 24 hours and 12 hours, respectively. Figure \ref{fig:acc} shows the spatial density of traffic accidents in each MSOA for 2021. \redfont{It demonstrates the spatial inequality in accident risk across all regions, and the non-grid spatial shapes provide a density measurement based on socio-demographic and size exposure effects.}
Figure \ref{fig:daily_accidents} illustrates the daily count of traffic accidents in London throughout 2021, along with a 7-day moving average to smooth the trend and highlight patterns over time.

For multiview urban configuration, we used POI data and road network data from Ordnance Survey through the Digimap platform\footnote{\url{https://digimap.edina.ac.uk/help/our-maps-and-data/os_overview/}}. Meteorological data was obtained from the MIDAS platform\footnote{\url{https://archive.ceda.ac.uk/}}, providing hourly weather information. Calendar information was automatically generated using the Python package \textit{‘holidays’}\footnote{\url{https://pypi.org/project/holidays/}}.

\subsection{Baselines}

For a comprehensive comparison, we choose four major types of model with in total 9 models as the baselines and SOTA models. The details are listed below. 

\begin{itemize}
\item \textbf{Traditional Models}
    \begin{itemize}
    \item \textbf{ARIMA \citep{box1976time}}: The Autoregressive Integrated Moving Average (ARIMA) is a classical model used for time series prediction. It predicts future accidents based on the average of historical accident risk as well as autoregressive patterns.
    \item \textbf{SVR \citep{drucker1996support}}: Support Vector Regression (SVR) is a variant of Support Vector Machine (SVM) designed for regression tasks. It predicts future sequences by training an optimal regression curve and allows accident data points to deviate from the regression curve to enhance generalisation capabilities.
    \end{itemize}
\item \textbf{Spatial/Temporal Deep Learning Models}
    \begin{itemize}
    \item \textbf{Hetero-Convlstm \citep{yuan2018hetero}}: The Convolutional Long Short-Term Memory (Hetero-ConvLSTM) network is the first model to address the spatial heterogeneity challenge in the large amount of accident data, while handling long-term dependencies.
    \item \textbf{GRU \citep{cho2014learning}}: The Gated Recurrent Unit (GRU) is a variant of RNNs with fewer parameters and a simpler structure compared to the traditional LSTM. It uses two gates, the update gate and the reset gate, to manage the flow of information.
    \item \textbf{GCN \citep{kipf2016semi}}: The Graph Convolutional Network (GCN) is a classical spatial learning model capable of aggregating feature information from surrounding locations layer by layer.
    \end{itemize}
\item \textbf{Spatiotemporal Graph Models}
    \begin{itemize}
    \item \textbf{STGCN \citep{yu2017spatio}}: The Spatiotemporal Graph Convolutional Network (STGCN) leverages spatial convolutions to aggregate feature information from adjacent geographical locations and gated convolutions to enhance temporal feature extraction, making it popular for spatiotemporal data mining.
    \item \textbf{DSTGCN \citep{yu2021deep}}: The Deep Spatio-Temporal Graph Convolutional Network (DSTGCN) integrates spatial learning layers and spatiotemporal learning layers to learn from static and dynamic data. Consider both the accident sequence and the influence of external factors, representing the state of the art in this category.
    \end{itemize}
\item \textbf{Spatiotemporal HyperGraph Models}
    \begin{itemize}
    \item \textbf{HyGCN \citep{wang2024hypergraph}}: The Hybrid Graph Convolutional Network (HyGCN) represents state-of-the-art work in hypergraphs-based urban computing. It captures higher-order relationships, achieving a spatial fusion of pairwise lower-order and higher-order relationship features in urban mobility flow predictions.
    \item \textbf{SST-DHL \citep{cui2024advancing}}: The Sparse Spatio-Temporal Dynamic Hypergraph Learning (SST-DHL) model represents the latest advancement in self-supervised accident prediction by addressing the sparsity issue with hypergraph. It employs a cross-regional dynamic hypergraph learning approach to identify global spatiotemporal dependencies of accdient data and introduces contrastive learning to provide supervisory signals, thereby enhancing model performance.
    \end{itemize}
\end{itemize}

\textbf{Experiment Configuration and Hyperparameter Settings:} The proposed SMA-Hyper model is implemented using PyTorch, and all training and testing are conducted on an NVIDIA A40 GPU. We use the Adam optimiser to optimise model parameters, with a learning rate set to 0.001. The batch size is adjusted to {4, 8, 16} based on GPU memory usage. The ratio of hyperedges to regions is set to 0.1, retaining the top 40 regions with the highest weights in each hyperedge. The embedding dimension is uniformly set to 32, and the number of attention heads is set to 8. To learn high-order complex spatiotemporal relationships, we stack 2 layers each of spatiotemporal encoders and convolutional decoders. In the loss function, we choose the coefficients as 0.001. The sequence length is set to 12 steps, with an 80\% training ratio.

\textbf{Evaluation Metrics:} We employ Root Mean Squared Error (RMSE) and Mean Absolute Error (MAE) to measure the accuracy of accident risk predictions, as well as Recall at top K (Rec@K) to evaluate the ability to identify high-risk regions wherereal accidents happen. For our evaluation, we choose $K = 20\%$. The formulas for multi-step predictions are as follows:

\begin{equation}
\text{RMSE} = \sqrt{\frac{1}{L} \sum_{t’=T+1}^{T+L} (y_{t’} - \hat{y}_{t’})^2}
\end{equation}

\begin{equation}
\text{MAE} = \frac{1}{L} \sum_{t’=T+1}^{T+L} |y_{t’} - \hat{y}_{t’}|
\end{equation}

\begin{equation}
\text{Rec@K} = \frac{1}{L} \sum_{t’=T+1}^{T+L} \frac{|R_{t’} \cap \hat{R}_{t’}|}{|R_{t’}|}
\end{equation}

where $y_{t’}$ and $\hat{y}_{t’}$ are the true and predicted risk scores of all districts in the time interval $t’$, respectively; and $ L$ is the duration of the prediction. $R_{t’}$ and $\hat{R}_{t’}$ are the sets of the top actual and predicted districts of the highest risk $K$, respectively. Given that accident risk data are zero-inflated with a large amount of nonrisk (zero) entries and is very left-skewed, lower RMSE values indicate more accurate predictions by heavily penalising larger errors, making RMSE particularly sensitive to extreme error values. Lower MAE values reflect more accurate predictions by providing a straightforward average error measure, which is less influenced by extreme values compared to RMSE, making it useful for understanding average prediction accuracy across all regions. Higher Rec@K values indicate the effectiveness of the model in identifying high-risk districts where real accidents occur.

\subsection{Results}

\subsubsection{Overall Performance}

To comprehensively evaluate the effectiveness of the proposed SMA-Hyper model, we conducted experiments on two temporal scales of the London accident dataset: 12-hour and 24-hour intervals. For these experiments, the model uses 12 input steps to generate multi-step predictions over 6 output steps. The sparsity of the data is indicated by the ratio of non-zero instances within the tensor of each dataset, reflecting the inherent challenges of predicting rare events. 

\textbf{MultiStep Overall Performance:} Table \ref{table1} summarises the performance metrics of the models, averaged over the multiple time horizons. This provides a clear comparison of the overall predictive accuracy of each model across multiple time horizons. Our SMA-Hyper model consistently demonstrates superior average performance compared to both traditional and state-of-the-art baseline models. Specifically, it achieves over 50\% improvement in regression metrics such as RMSE and MAE, and approximately 13.79\% and 13.96\% improvements in accurately identifying scenarios where accidents occurred within the predicted high-risk regions. In particular, Table \ref{table1a} shows that the SMA-Hyper model yields more accurate prediction results for the 12-hour interval compared to the 24-hour interval. However, the improvements are more pronounced in the 24-hour interval, which has a higher occurrence of accidents (6.01\% non-zero instances) as indicated in Table \ref{table1b}. The increased improvement in the 24-hour interval can be attributed to the higher volume of accidents, providing a richer accident scenario that allows the model to be sensitive to accident patterns.

However, despite the greater improvements, the 24-hour interval data show less accuracy compared to the 12-hour data. This is likely due to the coarser temporal resolution of the 24-hour interval, which aggregates accident data over longer periods, potentially smoothing out finer details that could be crucial for precise predictions. The 12-hour interval, with its finer temporal granularity, retains more detailed encoding information about accident occurrences, leading to higher accuracy in predictions. In particular, the adaptive graph learning component and the contrastive fusion of multi-urban views in both setwise graph and pairwise Hypergraph modules enable our model to effectively learn spatial heterogeneities and overcome spatial-temporal oversmoothing issues associated with sparsity. This results in better recall scores for accident identification(around 40\% in both temporal resolution datasets), demonstrating the robustness and effectiveness of the SMA-Hyper model in handling complex urban traffic accident prediction tasks with multi-urban views.

Moreover, in the baseline models, we observe an overall inferior performance with the 24-hour data compared to the corresponding 12-hour data, due to the fact that there are less data available for sufficient training with coarse-grained temporal levels. Despite this, the hypergraph-based models continue to lead the baselines in individual metrics. In particular, SST-DHL, the state-of-the-art hypergraph model for accident prediction, demonstrates a promising approach in predicting overall accident risk scores with lower errors in RMSE among all baselines. However, its performance is restricted by relying solely on self-supervised learning from historical accident data, lacking integration of urban configuration views \citep{cui2024advancing}.

The STGCN model shows the highest Recall performance compared to hypergraph-based baseline models. This superior Recall performance can be attributed to the effective use of spatial convolutions by STGCN, which aggregate feature information from adjacent geographical locations, and gated convolutions with enhanced temporal feature extraction. This dual approach allows STGCN to capture more relevant spatiotemporal patterns, making it more adept at identifying high-risk regions where actual accidents occur. In contrast, while hypergraph-based models like SST-DHL can capture higher-order relationships and offer robust predictions, their complexity might lead to overfitting, especially in sparse datasets. The intricate nature of hypergraphs can sometimes obscure the direct spatiotemporal correlations needed for accurate high-risk identification, thus affecting their Recall performance \citep{wang2024hypergraph}. It shows the need for the sparsity-considered hypergraph learning framework. 

\begin{figure}[htbp]
    \centering
    
    \begin{subfigure}[b]{\textwidth}
        \centering
        \includegraphics[width=\textwidth]{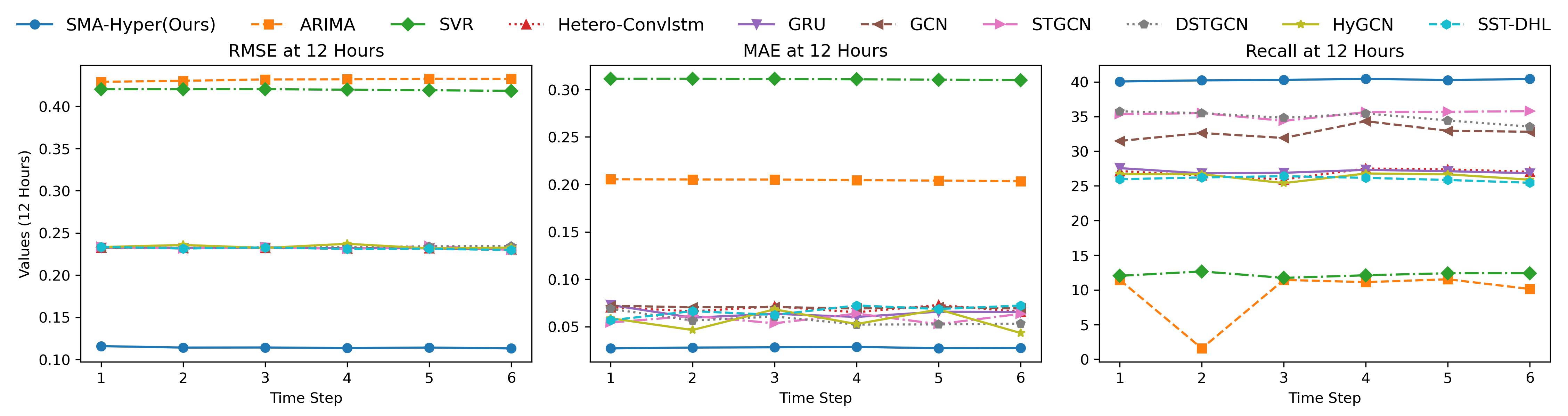}
        \caption{Model performance at 12 hours across all evaluation metrics}
        \label{fig:12hour_time}
    \end{subfigure}
    \hfill
    \vspace{0.12cm}
    \begin{subfigure}[b]{\textwidth}
        \centering
        \includegraphics[width=\textwidth]{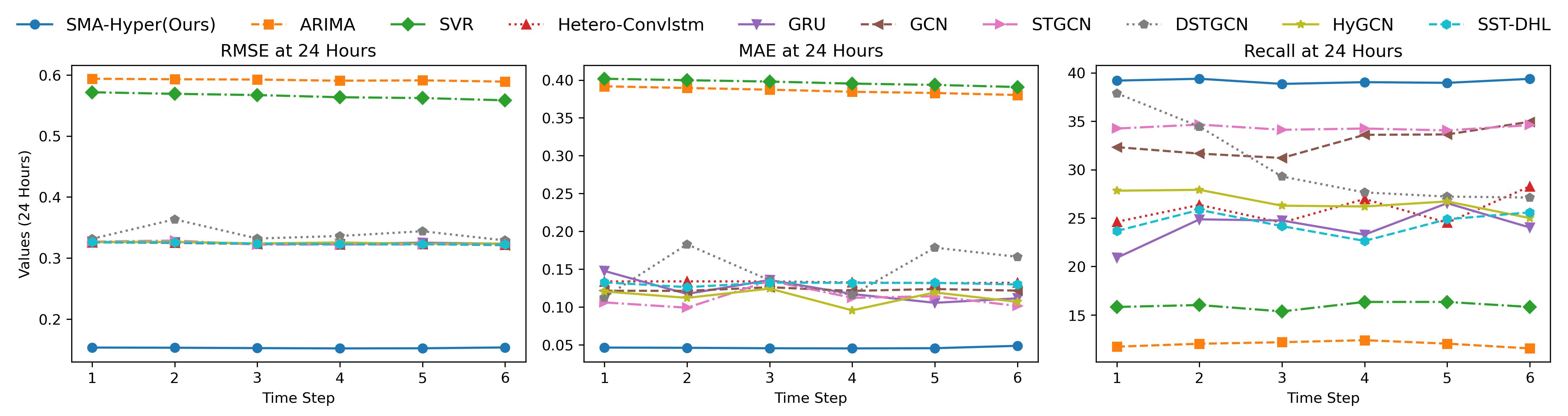}
        \caption{Model performance at 24 hours across all evaluation metrics}
        \label{fig:24hour_time}
    \end{subfigure}

    \caption{Step-wise performance on all evaluation metrics for different models.}
    \label{fig:timewise}
\end{figure}

\textbf{Timewise Stability Performance:} To gain a comprehensive understanding of our SMA-Hyper model’s consistent strengths in predicting the risks of traffic accidents under varying conditions, Figure \ref{fig:timewise} presents the timewise performance. Specifically, the 24-hour baseline models exhibit more fluctuations, as indicated by Figure \ref{fig:24hour_time}, due to limited training data. When comparing all the metrics across the four major types of models, the hypergraph-based models demonstrate relatively stable performance over time. Among these, the SMA-Hyper model shows the highest stability in Recall. This stability can be attributed to the high-order spatial relationships captured by hypergraphs, which enable the model to maintain consistent performance despite the variability in data. Thus, By effectively capturing complex spatiotemporal patterns in a reliable way, ~\M can better support the multi-steps management and planning of traffic safety interventions.

\begin{figure}[htbp]
    \centering
    \includegraphics[width=\textwidth]{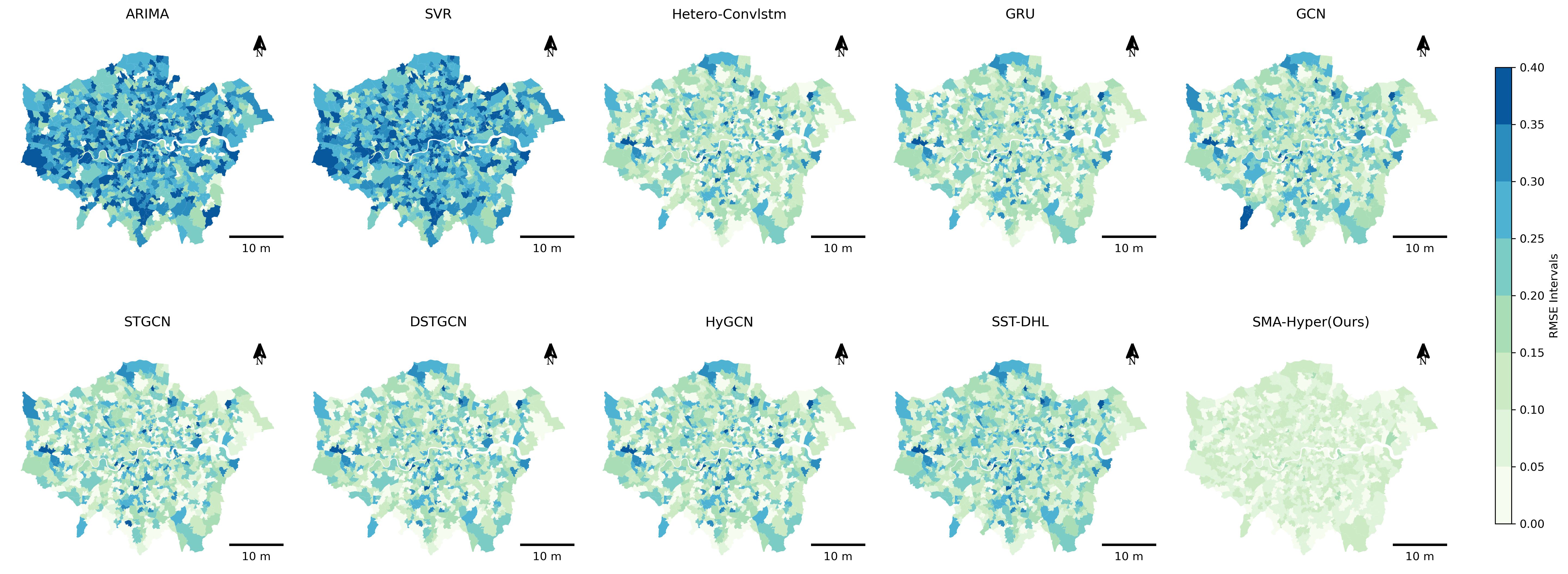}
    \caption{Prediction Error Visualization of different methods over MSOA regions in the entire urban space of London (12 Hours Results).}
    \label{fig:rmse_12}
\end{figure}

\textbf{Spatial Robustness Performance:} Figure \ref{fig:rmse_12} illustrates the spatial robustness performance of the SMA-Hyper model across various regions, encompassing non-risk, low-risk, and high-risk accident occurrences. The accident density map (Figure \ref{fig:acc}) reveals that accidents predominantly occur in central London and a few regions in outer London, surrounded by a large number of non-accident regions. The SMA-Hyper model integrates urban configurations as external supplementary information and transcends local spatial constraints by capturing high-order relational similarities among different regions. This approach enhances global fairness in encoding performance, mitigating issues such as overfitting or oversmoothing for outputs, particularly in critical locations.

Traditional graph neural networks (GNNs) are constrained in accident prediction primarily due to their reliance on pairwise relationships \citep{cui2024advancing}. These models aggregate information from immediate neighbours based on predefined spatial links, which may not adequately capture the complex, non-local interactions inherent in urban traffic systems \citep{li2024developing}. Given the highly localised nature of accident data, with a significant concentration in central regions and sparse occurrences in others, GNNs often struggle to generalise throughout the spatial landscape. This limitation is exacerbated by the zero-inflation issue, where the predominance of non-accident regions leads to oversmoothing and reduced model sensitivity to actual accident hotspots.

On the other hand, pure hypergraph models that focus solely on accident data are also constrained. While hypergraphs can capture higher-order relationships and provide a richer representation of complex interactions, models that rely exclusively on historical accident data lack the contextual information provided by urban configurations \citep{wang2024hypergraph}. This limitation reduces their ability to generalise and predict accidents in regions where historical data is sparse or non-existent. Additionally, self-supervised learning frameworks based solely on accident data may not fully leverage the spatial and urban heterogeneities that influence accident occurrences.

The SMA-Hyper model addresses these limitations by integrating both traditional graph-based approaches and hypergraph learning. By incorporating urban configuration data and capturing high-order spatial relationships through hypergraphs, the SMA-Hyper model effectively learns from the complex, multi-faceted nature of urban environments. This comprehensive approach allows the model to handle the inherent sparsity and variability in traffic accident data, leading to improved predictive accuracy and spatial fairness. Consequently, the SMA-Hyper model offers a more reliable and equitable foundation for urban traffic management and decision-making processes, supporting the development of targeted interventions and policies to enhance road safety.

\begin{table}[htpb]
    \small
    \centering
    \begin{subtable}{\textwidth}
        \centering
        \begin{tabular}{llccc}
        \toprule
        \hline
        \multirow{2}{*}{Model Class} & \multirow{2}{*}{Model} & \multicolumn{3}{c}{12 Hours (3.07\%)} \\
        \cmidrule(lr){3-5}
        & & RMSE & MAE & Recall@20\% \\
        \hline
        \multirow{2}{*}{\makecell[l]{Traditional Models}} 
        & ARIMA & 0.4315 & 0.2046 & 11.20 \\
        & SVR & 0.4198 & 0.3109 & 12.22 \\
        \hline
           \multirow{3}{*}{\makecell[l]{Spatial/Temporal Deep Learning Models}} 
        & Hetero-Convlstm & 0.2319 & 0.0686 & 26.91 \\
        & GRU & 0.2319 & 0.0645  & 27.08 \\
        & GCN & 0.2317 & 0.0703 & 32.68 \\
        \hline
        \multirow{2}{*}{\makecell[l]{Spatiotemporal Graph Models}} 
        & STGCN & 0.2322 & 0.0581 & \underline{35.40} \\
        & DSTGCN & 0.2331 & 0.0572 & 34.92 \\
        \hline
        \multirow{3}{*}{\makecell[l]{Spatiotemporal HyperGraph  Models}} 
        & HyGCN & 0.2337 & \underline{0.0562} & 26.36 \\
        & SST-DHL & \underline{0.2316} & 0.0664 & 26.01 \\
        & \M(Ours) & \textbf{0.1140} & \textbf{0.0277} & \textbf{40.28} \\
        \midrule
        & Improvements & \textbf{50.77\%}$\downarrow$ & \textbf{50.71\%}$\downarrow$   & \textbf{13.79\%}$\uparrow$  \\
        \hline
        \bottomrule
        \end{tabular}
        \caption{Performance comparison for 12 Hours time interval.}
        \label{table1a}
    \end{subtable}
    
    \bigskip
    
    \begin{subtable}{\textwidth}
        \centering
        \begin{tabular}{llccc}
        \toprule
        \hline
        \multirow{2}{*}{Model Class} & \multirow{2}{*}{Model} & \multicolumn{3}{c}{24 Hours (6.01\%)} \\
        \cmidrule(lr){3-5}
        & & RMSE & MAE & Recall@20\% \\
        \hline
        \multirow{2}{*}{\makecell[l]{Traditional Models}} 
        & ARIMA & 0.5918 & 0.3860 & 11.98 \\
        & SVR & 0.5654 & 0.3966 & 15.95 \\
        \hline
        \multirow{3}{*}{\makecell[l]{Spatial/Temporal Deep Learning Models}} 
        & Hetero-Convlstm & 0.3335 & 0.1329  & 25.88 \\
        & GRU & 0.3247 & 0.1224 & 24.06 \\
        & GCN & 0.3234 & 0.1226 & 32.90 \\
        \hline
        \multirow{2}{*}{\makecell[l]{Spatiotemporal Graph Models}} 
        & STGCN & 0.3242 & 0.1112 & \underline{34.32} \\
        & DSTGCN & 0.3394 & 0.1485  & 27.27 \\
        \hline
        \multirow{3}{*}{\makecell[l]{Spatiotemporal HyperGraph Models}} 
        & HyGCN & 0.3249 & 0.1136  & 26.66 \\
        & SST-DHL & \underline{0.3238} & \underline{0.1108}  & 24.47 \\
        & \M (Ours) & \textbf{0.1527} & \textbf{0.0462} & \textbf{39.11} \\
        \midrule
        & Improvements & \textbf{52.84\%}$\downarrow$ & \textbf{58.30\%}$\downarrow$  & \textbf{13.96\%}$\uparrow$ \\
        \hline
        \bottomrule
        \end{tabular}
        \caption{Performance comparison for 24 Hours time interval.}
        \label{table1b}
    \end{subtable}
    
    \caption{Performance comparison for 12 \& 24 Hours time interval. The bold font means the best performance, while the underline indicates the second best performance. The zero-inflation rate is highlighted in parentheses.}
    \label{table1}
\end{table}

\subsubsection{HyperEdge Examination and Analysis}

\begin{figure}[htbp]
    \centering
    
    \begin{subfigure}[b]{\textwidth}
        \centering
        \includegraphics[width=\textwidth]{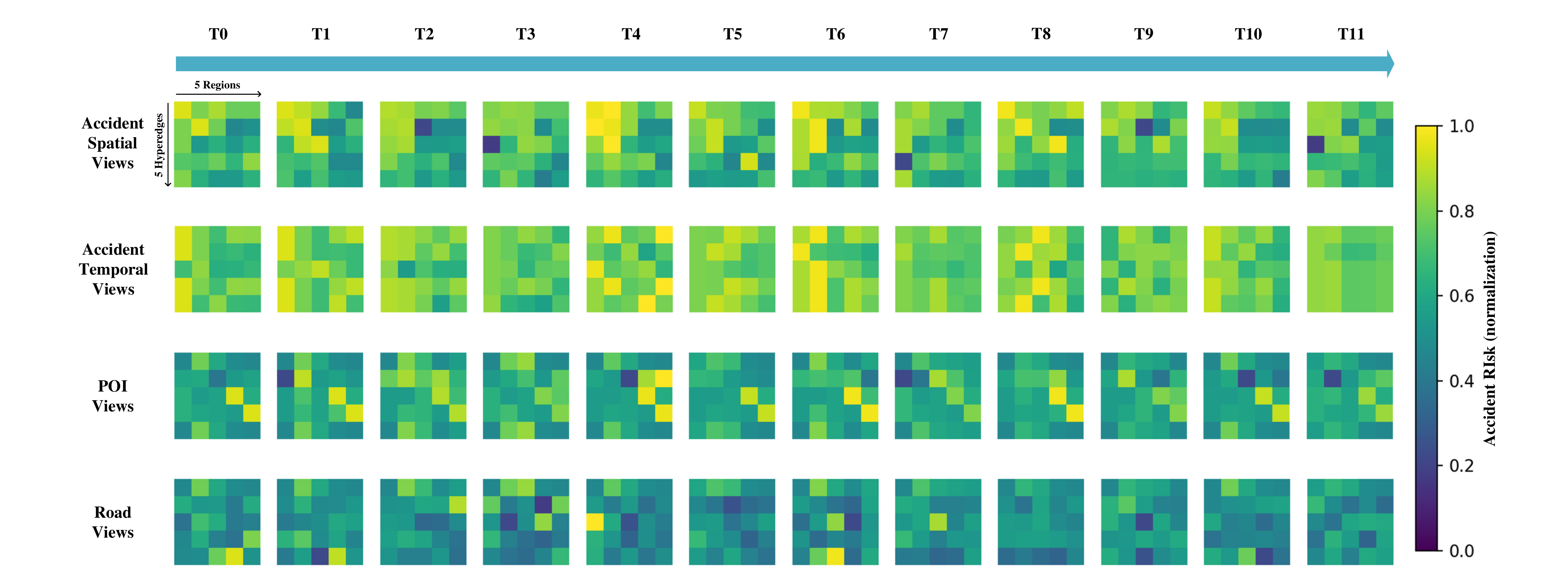}
        \caption{Temporal evaluation of the five most relevant regions to hyperedges. These maps highlight the temporal dynamics of hyperedges in different regions, providing insights into how certain areas are affected over time. The five regions on each hyperedge are expected to have similar accident risk scores, indicating a consistent pattern of hypergraph dynamics.}
        \label{fig:hyperedge_time}
    \end{subfigure}
    \hfill
    \begin{subfigure}[b]{\textwidth}
        \centering
        \includegraphics[width=\textwidth]{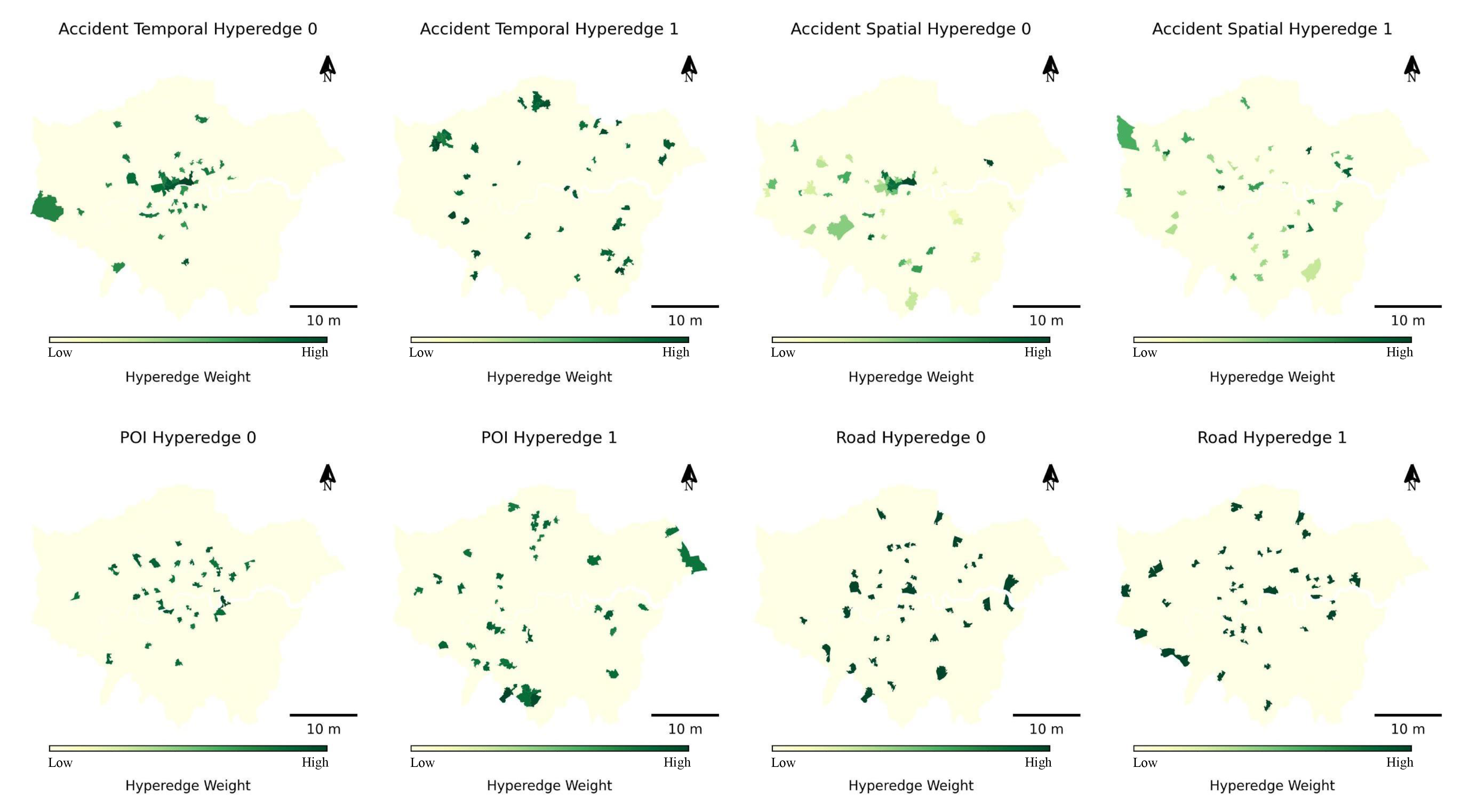}
        \caption{Spatial heat maps for the encoded relations of regions within sampled local hyperedges. These maps show the spatial distribution and intensity of hyperedges across different regions, illustrating how local factors influence hyperedge formation and relevance. It is expected that each hyperedge would capture the high-order cross-regional relations, reflecting complex spatial interactions.}
        \label{fig:hyperedge_spatial}
    \end{subfigure}
    
    \caption{Visualization of relevance weights of hyperedges from hypergraph learning. The left panel (a) provides a temporal evaluation of the five most relevant regions to hyperedges, while the right panel (b) presents spatial heat maps of regions to local hyperedges. }
    \label{fig:hyperedge_comparison}
\end{figure}

Figure \ref{fig:hyperedge_comparison} presents a detailed analysis of hyperedges, utilizing multiple urban views to understand how different factors contribute to accident risks. To maintain consistency in the analysis, min–max normalization is applied to all data, allowing us to dynamically capture complex, high-order correlations of traffic accidents occurring both near and far among all city regions \citep{cui2024advancing}.

Figure \ref{fig:hyperedge_time} illustrates the temporal dynamics of each hyperedge in the five most relevant regions, providing a clear representation of how the dynamic hypergraph captures higher-order correlations with time variation. Each view captures different aspects of the urban environment, and the shades of color reveal traffic accident risk patterns in different views. The similar colors in each hypergraph clearly demonstrate that our SMA-Hyper model can capture the global dependence of different geographical regions in various urban configurations with similar traffic accident scenarios. Furthermore, the dynamic hyperedges consistently maintain similar global dependencies over input time, as illustrated by the nuanced changes in colors in the correlation matrix as time shifts.

Figure \ref{fig:hyperedge_spatial} provides spatial heat maps for the encoded relations of regions within the sampled local hyperedges. These maps illustrate the complex spatial distribution and intensity of hyperedges across different regions, showing how local factors influence the formation and relevance of hyperedges. Compared to the ground truth density map in Figure \ref{fig:acc}, the sampled hyperedges effectively capture similar-density accident regions within each hyperedge, demonstrating the model’s ability to identify critical areas. For example, as depicted by hyperedge 0 in the accident views, this ability to capture regions with similar dense accident occurrence patterns, even if they are distant from each other, underscores the robustness and effectiveness of the SMA-Hyper model.

\subsubsection{Ablation Studies with Model Configurations}

\begin{figure}[htbp]
    \centering  
    \begin{subfigure}[b]{0.7\textwidth}
        \centering
        \includegraphics[width=\textwidth]{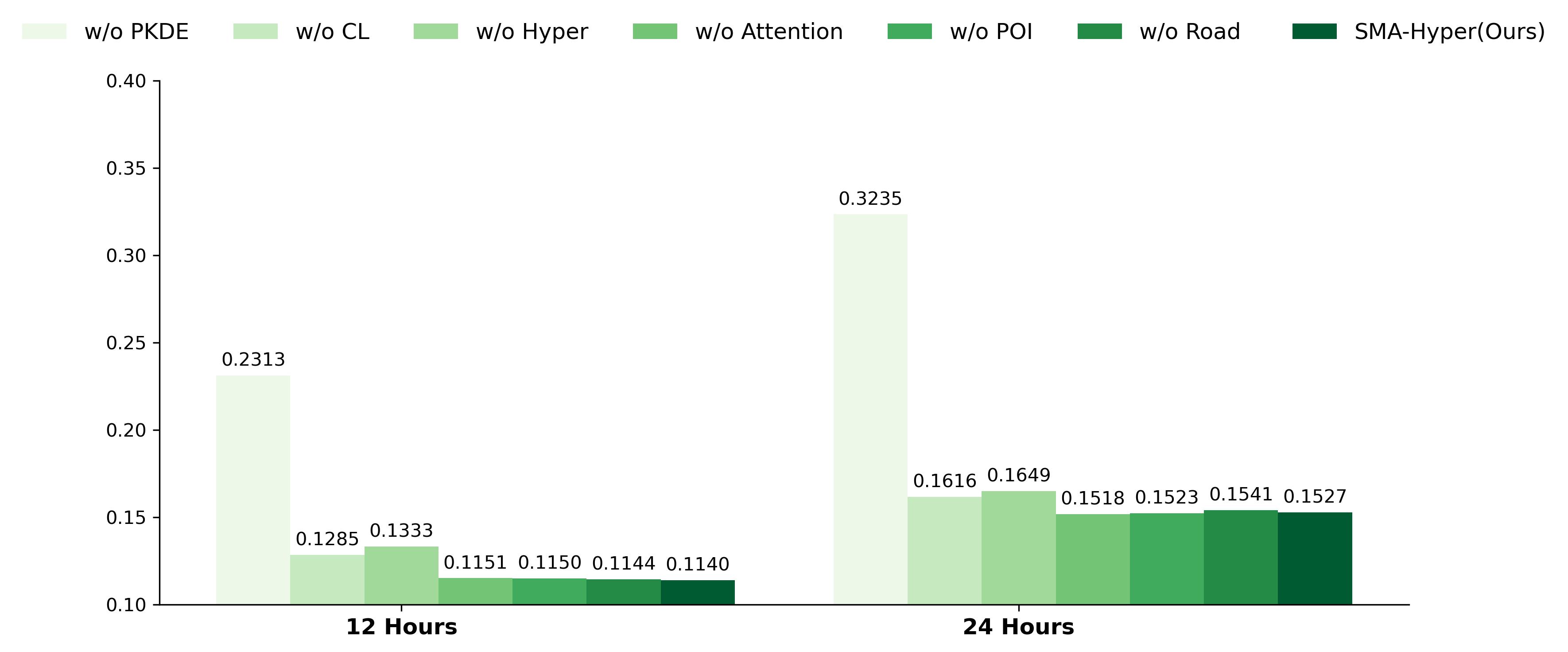}
        \caption{RMSE at 12 Hours and 24 Hours.}
        \label{fig:ab_rmse}
    \end{subfigure}
    \begin{subfigure}[b]{0.7\textwidth}
        \centering
        \includegraphics[width=\textwidth]{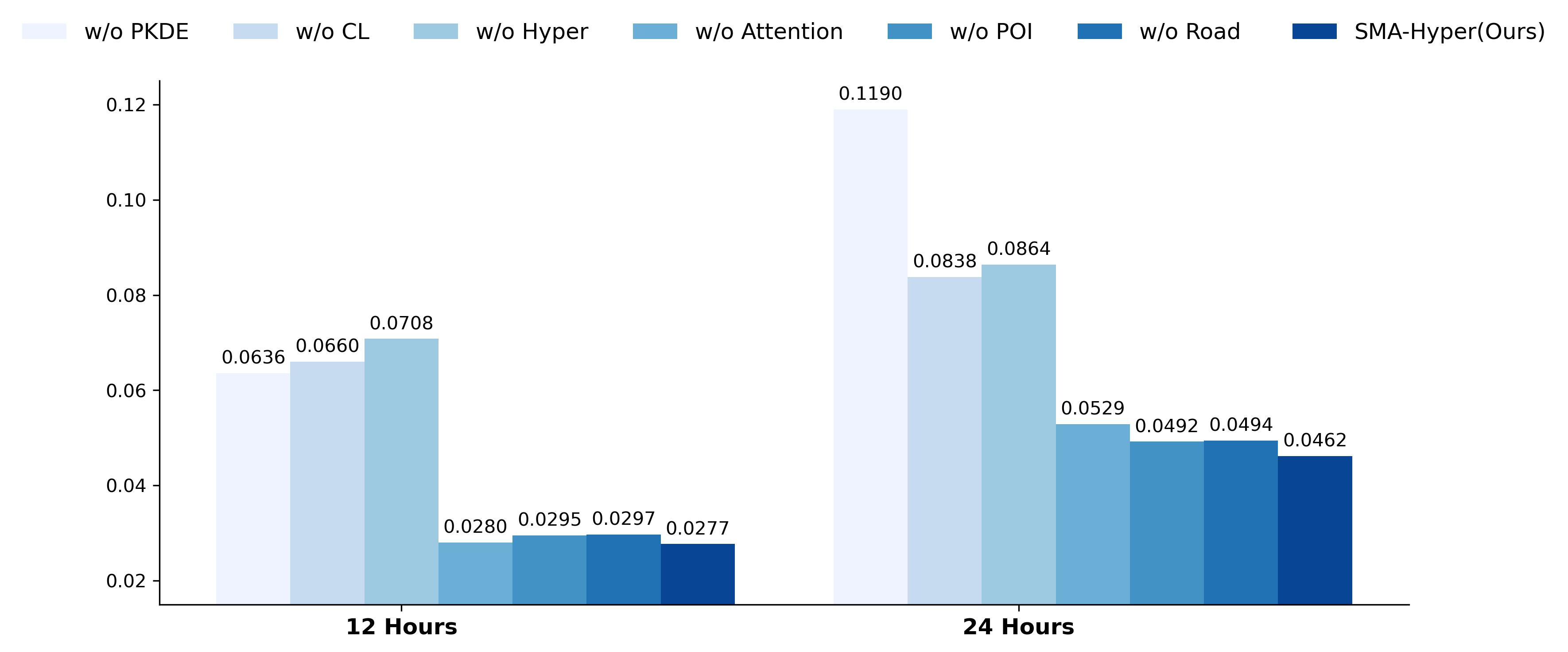}
        \caption{MAE at 12 Hours and 24 Hours.}
        \label{fig:mae}
    \end{subfigure}
    
    \begin{subfigure}[b]{0.7\textwidth}
        \centering
        \includegraphics[width=\textwidth]{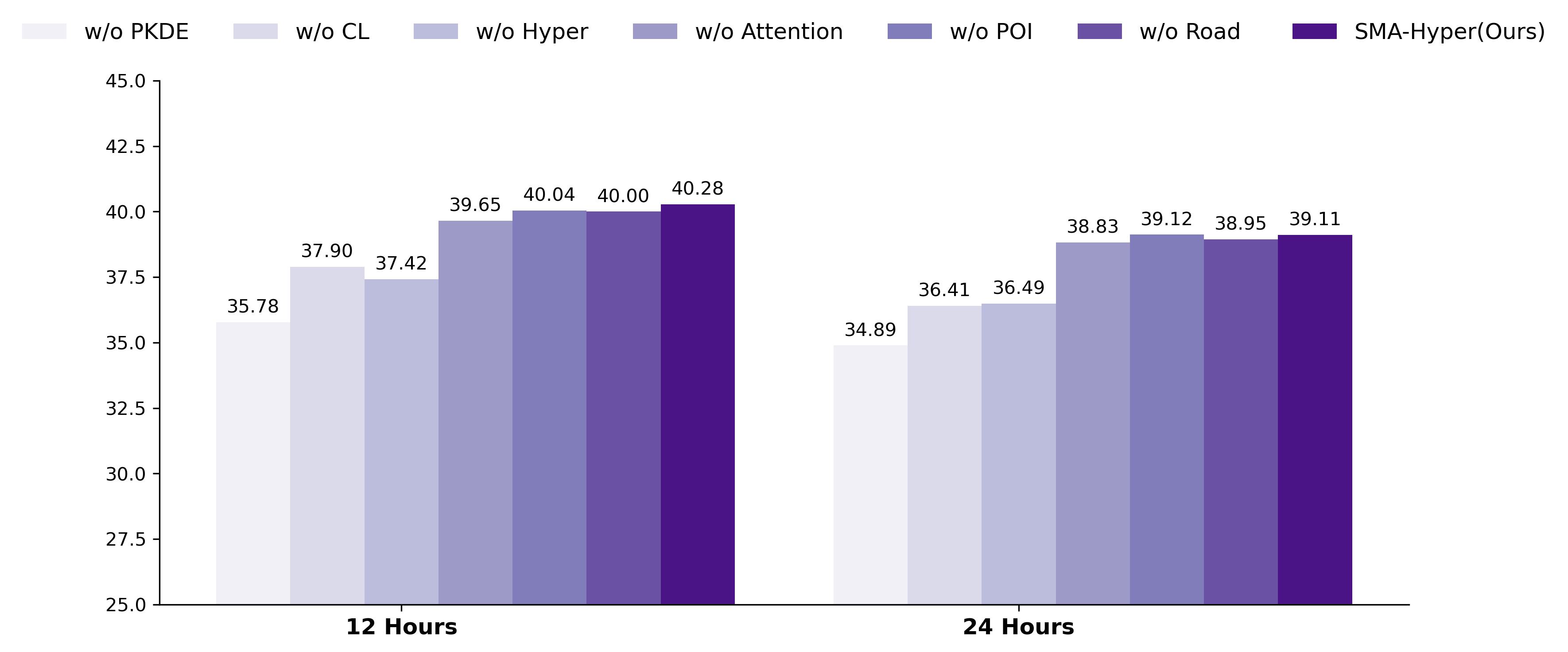}
        \caption{Recall at 12 Hours and 24 Hours.}
        \label{fig:recall}
    \end{subfigure}
    
    \caption{Performance comparison of different model configurations evaluated using MAE, RMSE, and Recall metrics at 12-hour and 24-hour intervals.}
    \label{fig:performance_comparison}
\end{figure}

To evaluate the impact of various components within our framework, including PKDE for mitigating imbalanced accident labels, contrastive learning, attention mechanisms, and hypergraph learning, an ablation study was conducted. This study aimed to evaluate the contributions of these components to the predictive precision of urban traffic accidents while integrating two urban configurations (POI and Roads) for comprehensive node representation. Each variable was individually added, while all other modules were kept constant, as shown in Figure \ref{fig:performance_comparison}.

In general, PKDE demonstrated the most significant positive impact on the improvement of the SMA-Hyper model, resulting in substantial reductions across all metrics, particularly with an increase in recall of more than 10\%. It is important to note that the baseline models referred to in Table \ref{table1} were all tested without the PKDE strategy. However, even when this strategy is eliminated, the SMA-Hyper model consistently outperforms these baseline models.

Regarding the ablation performance of cCL, hypergraph learning, and the attention mechanism, three key elements of our framework, the absence of hypergraph learning resulted in the most significant decrease in performance. Specifically, the reduction ratios ranged from 7.98\% to 9.47\% in RMSE, 155.59\% to 87.00\% in MAE and 7.1\% to 7.7\% in Recall. Furthermore, the 24-hour data set, with its limited training data, showed greater sensitivity to all framework modules, further underscoring the importance of these components in our model.

For the urban view, incorporating the POI and Road data resulted in improved performance, although the improvement was modest. However, the Road view played a more significant role in contributing to better results. In particular, the inclusion of road information results in an improvement in MAE, highlighting the importance of road characteristics in accurately predicting accident risks, which also corresponds to previous studies \citep{mannering2014analytic,wang2021traffic,yang2022predicting}. This is likely due to the direct influence that the types, conditions, and infrastructure of roads have on traffic safety, making them critical factors in the predictive capabilities of the model.

\subsubsection{Sensitivity studies for hyperperameter}

We evaluated the model using different hyperparameters, including the loss parameter $\lambda_{1}$, the length of the input sequence, the training ratio, and the number of GCN layers, to justify our choices in corresponding settings. The experiments are carried out on 12-hour and 24-hour intervals, and performance is assessed using RMSE, MAE, and Recall metrics. The results are summarised in Table \ref{tab:ablation}. 

The parameter $\lambda_{1}$ controls the ratio of cross-graph and hypergraph contrastive learning, achieving balanced performance at 0.1. Although the RMSE is slightly lower for the 24-hour prediction at $\lambda_{1} = 1$, metrics such as MAE and Recall are more sensitive to accurate and precise predictions. The value of $\lambda_{1} = 0.1$ achieves superior results in these metrics compared to $\lambda_{1} = 1$, indicating that this setting helps the model better identify high-risk regions and maintain lower prediction errors. This is crucial for practical applications where accurately pinpointing accident-prone areas is more important than marginal improvements in RMSE.

A similar situation occurs with the GCN layers. We chose 2 layers as the optimal parameter, although 3 layers show slightly lower RMSE for the 24-hour interval. Increasing the number of GCN layers can lead to over-smoothing, where node representations become indistinguishable due to excessive aggregation of information. This degrades the ability of the model to distinguish between different nodes, leading to poorer performance on metrics such as MAE and Recall, which are sensitive to precise predictions. Furthermore, adding more layers increases the complexity of the model, which can lead to overfitting, especially when training data is sparse, such as in the case of 24-hour intervals. Overfitting can result in the model capturing noise rather than meaningful patterns, negatively affecting generalisation performance.

For sequence length and training ratios, these hyperparameters are all about the quality of training data. Although higher training ratios show better results, longer input sequence lengths are not always optimal. Longer sequences can introduce more noises and uncertainties into the training process, potentially degrading model performance\citep{trirat2023mg}.

\begin{table}[htbp]
    \centering

    \begin{subtable}[t]{\textwidth}
        \centering
        \begin{tabular}{p{2cm}<{\centering}|p{2.5cm}<{\centering}p{2.5cm}<{\centering}p{2.5cm}<{\centering}}
            \toprule
            \hline
            $\lambda_{1}$ &  \multicolumn{3}{c}{12 Hour / 24 Hours} \\
            \midrule
                &  RMSE & MAE & Recall (\%) \\
            0.001 & 0.1302/0.1648 & 0.0667/0.0849 & 37.55/36.33\\
            0.01 & 0.1164/0.1604 & 0.0304/0.0857 & 37.74/37.17 \\
            0.1 & \textbf{0.1140}/0.1527 & \textbf{0.0277/0.0462} & \textbf{40.28/39.11} \\
            1 & 0.1143/\textbf{0.1525} & 0.0288/0.0476 & 39.85/38.49 \\
            \hline
            \bottomrule
        \end{tabular}
        \caption{Performance on different tests of $\lambda_{1}$ as loss parameters.}
        \label{tab:lambda}
    \end{subtable}
    \hfill
    \vspace{0.3cm}
    \begin{subtable}[t]{\textwidth}
        \centering
        \begin{tabular}{p{2cm}<{\centering}|p{2.5cm}<{\centering}p{2.5cm}<{\centering}p{2.5cm}<{\centering}}
            \toprule
            \hline
            Sequence \\ Length &  \multicolumn{3}{c}{12 Hour / 24 Hours} \\
            \midrule
                &  RMSE & MAE & Recall (\%) \\
            8 & 0.1142/0.1531 & 0.0293/0.0488 & 39.02/37.00\\
            10 & 0.1145/0.1530 & 0.0288/0.0518 & 39.63/37.77 \\
            12 & \textbf{0.1140/0.1527} & \textbf{0.0277/0.0462} & \textbf{40.28/39.11} \\
            14 & 0.1143/0.1529 & 0.0284/0.0501 & 39.69/38.41 \\
            16 & 0.1145/0.1533 & 0.0279/0.0505 & 40.09/38.61 \\
            \hline
            \bottomrule
        \end{tabular}
        \caption{Performance on varying input sequence length.}
        \label{tab:length}
    \end{subtable}
    \hfill
    \vspace{0.3cm}
    \begin{subtable}[t]{\textwidth}
        \centering
        \begin{tabular}{p{2cm}<{\centering}|p{2.5cm}<{\centering}p{2.5cm}<{\centering}p{2.5cm}<{\centering}}
            \toprule
            \hline
            Training \\ Ratio &  \multicolumn{3}{c}{12 Hour / 24 Hours} \\
            \midrule
                &  RMSE & MAE & Recall (\%) \\
            0.5 & 0.1161/0.1533 & 0.0305/0.0477 & 38.22/38.83\\
            0.6 & 0.1154/0.1528 & 0.0294/0.0499 & 39.74/38.93 \\
            0.7 & 0.1172/0.1544 & 0.0291/0.0495 & 39.94/38.81 \\
            0.8 & \textbf{0.1140/0.1527} & \textbf{0.0277/0.0462} & \textbf{40.28/39.11} \\
            \hline
            \bottomrule
        \end{tabular}
        \caption{Performance on training ratio.}
        \label{tab:train_ratio}
    \end{subtable}
    \hfill
    \vspace{0.3cm}
    \begin{subtable}[t]{\textwidth}
        \centering
        \begin{tabular}{p{2cm}<{\centering}|p{2.5cm}<{\centering}p{2.5cm}<{\centering}p{2.5cm}<{\centering}}
            \toprule
            \hline
            GCNs \\ Layers &  \multicolumn{3}{c}{12 Hour / 24 Hours} \\
            \midrule
                &  RMSE & MAE & Recall (\%) \\
            1 & 0.1306/0.1636 & 0.0655/0.0842 & 37.89/36.93\\
            2 & \textbf{0.1140}/0.1527 & \textbf{0.0277/0.0462} & \textbf{40.28/39.11} \\
            3 & 0.1145/\textbf{0.1502} & 0.0283/0.0488 & 40.07/38.89 \\
            4 & 0.1149/0.1539 & 0.0284/0.0478 & 39.92/38.90 \\
            5 & 0.1145/0.1508 & 0.0286/0.0499 & 40.07/38.91 \\
            \hline
            \bottomrule
        \end{tabular}
        \caption{Performance on the number of GCNs layers.}
        \label{tab:layer_ab}
    \end{subtable}
     \caption{Ablation study: Comparison of performance on different tests of $\lambda_{1}$ as loss parameters, varying input sequence length, training ratio and GCNs layers for 12-hour and 24-hour intervals. Each table presents the RMSE, MAE, and Recall metrics.}
    \label{tab:ablation}
\end{table}

\section{Conclusion}
\label{sec:con}

The SMA-Hyper framework consistently demonstrates superior and robust performance in multistep urban traffic accident prediction by effectively addressing the challenges posed by data sparsity and the need for precise identification of high-risk areas. It integrates adaptive graph learning and employs both setwise and pairwise contrastive graph representation learning techniques to dynamically capture spatial heterogeneities and homogeneities. Additionally, with the help of PKDE, the framework overcomes the limitations of traditional and state-of-the-art models. By leveraging hypergraph learning, the SMA-Hyper model effectively captures high-order spatiotemporal dynamics, offering a comprehensive understanding of the complex interactions between various urban factors. Incorporating diverse urban configurations, such as POI and road network data, the model achieves a more feasible and explainable representation of the urban environment. This multiview fusion approach enables the model to adapt to the dynamic nature of urban traffic patterns, thereby improving predictive accuracy across different temporal scales.

Extensive experiments on the London traffic accident dataset reveal that the SMA-Hyper model significantly improves key performance metrics, including RMSE, MAE, and Recall. The model’s ability to maintain high performance under varying temporal conditions and data sparsity underscores its robustness and reliability for practical applications in urban traffic management.

Despite its strengths, the SMA-Hyper model has some limitations. The complexity of the model increases computational requirements, which may be a constraint in real-time applications. Furthermore, while the model integrates multiple urban views, it may still miss certain contextual nuances that could further enhance prediction accuracy. Future work could explore the inclusion of more diverse data sources and the optimisation of computational efficiency.

Overall, the SMA-Hyper model represents a significant advancement in multistep traffic accident prediction, providing valuable insights and a solid foundation for the development of targeted interventions and policies aimed at enhancing road safety in urban environments.

\section{CRediT authorship contribution statement}

\section{Declaration of Competing Interest}

The authors declare that they have no known competing financial interests or personal relationships that could have appeared to influence the work reported in this paper.

\section{Data availability}

Data could be downloaded publicly. 

\section{Acknowledgments}

The authors thank Dr. Lu Yin, Mr. Zheyan Qu, and the anonymous referees for their insightful comments.

\clearpage
\bibliographystyle{apalike} 
\bibliography{ref}

\end{document}